\pdfoutput=1

\documentclass[11pt]{article}

\usepackage[]{ACL2023}

\usepackage{times}
\usepackage{latexsym}
\usepackage{graphicx}
\usepackage{amsmath}
\DeclareMathOperator*{\argmax}{argmax}
\DeclareMathOperator*{\argmin}{argmin}
\usepackage{booktabs}
\usepackage{multirow}
\usepackage{subcaption}
\usepackage{arydshln}
\usepackage{soul}
\usepackage{algorithm}
\usepackage{algpseudocode}
\usepackage{rotating}
\usepackage{graphicx}
\usepackage{booktabs}
\definecolor{lightblue}{rgb}{0.68, 0.85, 0.9}
\usepackage{comment}

\usepackage[T1]{fontenc}

\usepackage[utf8]{inputenc}

\usepackage{microtype}

\usepackage{inconsolata}

\definecolor{todo}{rgb}{1,0.5,0}
\usepackage{enumitem}

\title{
Re3: A Holistic Framework and Dataset\\for Modeling Collaborative Document Revision
}

\author{Qian Ruan, Ilia Kuznetsov, Iryna Gurevych  \\
        Ubiquitous Knowledge Processing Lab (UKP Lab)\\
        Department of Computer Science and Hessian Center for AI (hessian.AI)\\
        Technical University of Darmstadt \\
  \texttt{www.ukp.tu-darmstadt.de}}

\begin{document}
\captionsetup[table]{skip=5pt}
\captionsetup[figure]{skip=5pt}
\maketitle
\begin{abstract}
Collaborative review and revision of textual documents is the core of knowledge work and a promising target for empirical analysis and NLP assistance. Yet, a holistic framework that would allow modeling complex relationships between document revisions, reviews and author responses is lacking. 
To address this gap, we introduce Re3, a framework for joint analysis of collaborative document revision. We instantiate this framework in the scholarly domain, and present Re3-Sci, a large corpus of aligned scientific paper revisions manually labeled according to their action and intent, and supplemented with the respective peer reviews and human-written edit summaries. We use the new data to provide first empirical insights into collaborative document revision in the academic domain, and to assess the capabilities of state-of-the-art LLMs at automating edit analysis and facilitating text-based collaboration. 
We make our annotation environment and protocols, the resulting data and experimental code publicly available.\footnote{\url{https://github.com/UKPLab/re3}}
\end{abstract}

\section{Introduction}
Textual documents are a key medium of information exchange in the modern world. These documents often result from a collaboration of multiple individuals. 
The typical process of collaborative text production involves iterations of drafting, getting feedback (\textit{\textbf{re}views}), executing \textit{\textbf{re}visions}, and providing \textit{\textbf{re}sponses} that outline the implemented changes, serving as a vital element in facilitating effective communication \citep{ape, f1000rd}. 
\begin{figure}[ht]
  \centering
\includegraphics[width=0.39\textwidth]{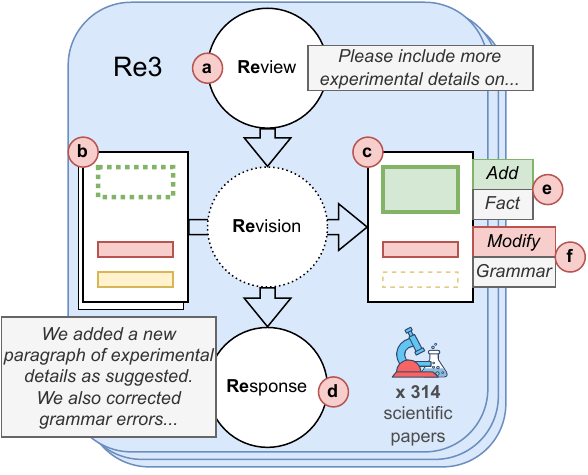}
    \caption{Re3 offers a holistic framework for studying the relationships between reviews (a), revisions (b-c) and responses (d) in text-based collaboration. It is instantiated in the Re3-Sci dataset that covers all edits in 314 full-length scientific publications manually labeled with edit action and intent (e) on different granularity levels, along with reviews that trigger edits and manually curated responses that summarize all edits made including self-initiated ones (f).}
    \label{fig:re3_procedure}
\end{figure}
Despite the importance of collaborative text revision and its high potential for NLP applications, we are missing a framework that formally describes this review-revision-response procedure grounded in real-world data.
While prior work in NLP has studied 
relationships between original and revised documents \citep{IteraTeR,arxivedits}, reviews and original documents \citep{nlpeer}, reviews and revisions \citep{f1000rd, aries}, and reviews and responses \citep{gao-rebuttal, ape} -- no prior frameworks allow jointly modeling all three components of text-based collaboration. Yet, such joint modeling is important as it provides deeper insights into the processes involved in text work, and opens new opportunities for NLP applications. Important tasks that involve reviews, revisions and responses such as \emph{edit summarization} thus remain underexplored.

Comprehensive analysis of document-level revisions poses additional challenges. Contrary to sentence-level analysis, hierarchically structured documents \citep{histruct} bring distinct levels of granularity into editing.
Individuals execute revisions at various granularity levels, with a range of actions and a spectrum of intents, reflecting the \textit{what}, \textit{how}, and \textit{why} of the revisions (Figure \ref{fig:re3_procedure} and §\ref{subsec:rev_dim}). 
Realistic modeling of document revision in text-based collaboration thus requires datasets and annotations that encompass the \textbf{\textit{entire document context}}, incorporating \textbf{\textit{all edits}} made across various levels of \textbf{\textit{granularity}}, and providing qualitative labels for both \textbf{\textit{action}} and \textbf{\textit{intent}}. We further term this kind of analysis as \textbf{full-scope} modeling of document revision.
Prior research in NLP has primarily studied sentence-level edits while neglecting the broader document context \citep{wpqac, wikieditintent-yang}, variations in granularity \cite{IteraTeR, argrewrite2}, and the underlying intent behind the edits \citep{newsedits, arxivedits}. There is thus a gap in both methodologies and datasets for creating and analyzing full-scope annotations of document revisions, limiting our grasp of the intricate nature of the editing process.

To close this gap and enable a comprehensive study of text-based collaboration in NLP, we introduce \textbf{Re3}: the first holistic framework for modeling review, revision and response in collaborative writing (§\ref{sec:framework}). We instantiate our framework in the scholarly domain and create \textbf{Re3-Sci}, the first large-scale human-annotated dataset that comprises 11.6k full-scope revision annotations for over 300 revised documents with substantial Inter-Annotator Agreement (IAA), as well as cross-document connections between reviews, revisions and responses (§\ref{sec:corpus}).
Our framework and dataset, for the first time, enable large-scale empirical investigation of collaborative document revision, including edit localization and clustering within documents, edit mechanisms and motivations inferred through action and intent labels, and the impact of review requests (§\ref{sec:analysis}).
Manually analyzing the complex relationships between reviews, revisions and responses is costly, and constitutes a promising NLP automation target. Facilitated by our data, we present a first exploration of the capability of large language models (LLMs) to address novel revision assistance tasks,  such as review request extraction, revision alignment, edit intent classification and document edit summarization (§\ref{sec:models}).
Our work thus makes four key \textbf{contributions}: 
\begin{itemize}[itemsep=-1pt, parsep=-1pt]
\item A holistic framework for studying document revisions and associated interactions in collaborative writing, including label taxonomy and robust annotation methodology;
\item A high-quality large-scale dataset that instantiates the framework in the domain of academic writing and peer review;
\item An in-depth analysis of human editing behavior in the scholarly domain;
\item Extensive experiments in automation with LLMs on four NLP tasks: review request extraction, revision alignment, edit intent classification and document edit summarization.
\end{itemize}
Our work paves the path towards comprehensive study of NLP for text-based collaboration in the scholarly domain and beyond.

\section{Related Work}
\label{subsec:related_work}
\begin{table}[ht]
\tabcolsep=0.06cm
\fontsize{9}{9}
\selectfont
\tabcolsep=0.09cm
\renewcommand{\arraystretch}{1.2} 
\begin{subtable}[ht]{0.48\textwidth}
       \centering
        \begin{tabular}{lllllll} \toprule
       &length &edits &
       full-scope 
       &align &intent \\ \hline
       \begin{tabular}[c]{@{}l@{}}IteraTeR  \citeyearpar{IteraTeR} \end{tabular}& 197 & 7* & no &4k*&4k*\\ 
       \begin{tabular}[c]{@{}l@{}}ArgRewrite \citeyearpar{argrewrite2}  \end{tabular}&582 & 19 & no & 3.2k & 3.2k\\
       \begin{tabular}[c]{@{}l@{}}arXivEdits \citeyearpar{arxivedits} \end{tabular} & 3,916 & 17 & no &
       \textbf{13k} & 1k \\
        \textbf{Re3-Sci (ours)} &
  \textbf{5,033} & \textbf{37} & \textbf{yes} &
  \textbf{11.6k} & \textbf{11.6k} \\ \bottomrule
       \end{tabular}
       \caption{Comparison of human-annotated document revision datasets.
       Presented are document length (words),  average sentence edits per document, presence of full-scope revision annotations, and data size, i.e., count of aligned and labeled sentence edits. * refers to subsentence edits as only such annotations are available. Re3-Sci is the first large-scale corpus with full-scope annotations of edit alignments, actions, and intents across multiple granularity levels in the entire document. }
       \label{tab:doc_rev_datasets}
\end{subtable}
\begin{subtable}[ht]{0.48\textwidth}
    \centering
        \begin{tabular}{lllll} \toprule
        &\begin{tabular}[c]{@{}l@{}} full-scope \end{tabular}   
       &\begin{tabular}[c]{@{}l@{}} review-\\revision \end{tabular} &\begin{tabular}[c]{@{}l@{}} revision-\\response \end{tabular}  &\begin{tabular}[c]{@{}l@{}} review-\\response \end{tabular} \\\hline 
       F1000RD \citeyearpar{f1000rd}&  no &yes&no&no\\
       NLPeer \citeyearpar{nlpeer} & no &yes &no&no \\
       ARIES \citeyearpar{aries}& no &yes&no&no\\
    \textbf{Re3-Sci (ours)} 
    & \textbf{yes} & 
  \textbf{yes} & \textbf{yes} & \textbf{yes} \\ \bottomrule
       \end{tabular}
       \caption{Comparison of review-revision-response datasets.
     Presented are presence of full-scope revision annotations, and interactions between the documents. Our work is the first to cover the entire review-revision-response procedure with full-scope annotations.}
       \label{tab:re3_datasets}
\end{subtable}
\caption{Related datasets comparison.}
\end{table}
\noindent\textbf{Document revision datasets.} 
Research on text revision originates in studies on Wikipedia \citep{wpqac, wikieditintent-yang, wikiatomicedits} and academic writing \cite{tan-lee-2014, xue-hwa-2014-redundancy}, which offer partial sentence-based annotations, neglecting the document context. 
Recent works have expanded the analysis to news articles \cite{newsedits}, student essays \citep{argrewrite,argrewrite2}, and scientific papers \citep{IteraTeR, arxivedits}.  
However, some focus mainly on revision alignment yet overlook the underlying intents \citep{newsedits, arxivedits}. Others restrict to specific sections or short texts, limiting analysis to sentence level \citep{IteraTeR, argrewrite, argrewrite2}. 
In this work, we introduce Re3-Sci, the first large-scale corpus with full-scope annotations of edit alignments, actions, and intents across multiple granularity levels in the entire document (Table \ref{tab:doc_rev_datasets}).

\noindent\textbf{Collaborative revision in peer review.}
Scholarly peer review is an essential example of collaborative text work in the academic domain. Open peer review provides an excellent opportunity to study the review-revision-response procedure. Prior works in NLP for peer review investigate argument mining-driven review analysis \citep{hua-etal-2019-argument, Fromm2020ArgumentMD} and the interplay between reviews and argumentative rebuttals \citep{gao-rebuttal, ape, ape2, disapere}, among others. 
Only a few studies and datasets investigate revision requests in peer reviews and their connection to the original texts \citep{nlpeer}, or to the actual revisions \citep{f1000rd, aries}. 
However, these do not provide full-scope annotations with qualitative labels and neglect self-initiated revisions not prompted by reviewers.
Our work is the first to cover the entire review-revision-response procedure with full-scope annotations in the context of scholarly publishing and peer review (Table \ref{tab:re3_datasets}).

\section{The Re3 Framework} 
\label{sec:framework} 
The Re3 framework builds upon the recently introduced intertextual model by \citet{f1000rd}. In particular, we represent documents as graphs that preserve document structure, allowing us to work on different levels of granularity, and treat cross-document relations as edges between the corresponding document graphs.
\begin{figure}[ht]
  \centering
    \includegraphics[width=0.47\textwidth]{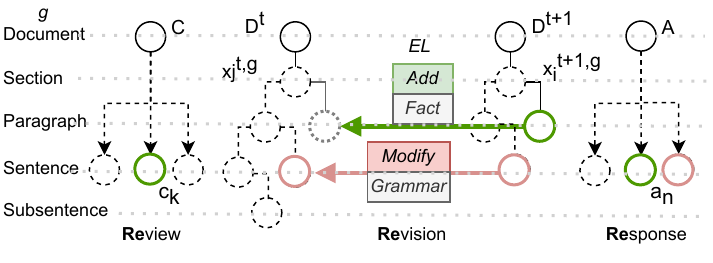}
    \caption{Illustration of the Re3 framework. Document revision, review and author response are represented as graphs, preserving structure and granularity through nodes and treating cross-document relations as edges. Refer to §\ref{subsec:model_term} for notation definitions.}
    \label{fig:re3}
\end{figure}
 
\subsection{Model and Terminology} 
\label{subsec:model_term}
As shown in Figure \ref{fig:re3}, we conceptualize the review-revision-response procedure as a set of interactions among four document types - the original document $D^t$, the revised document $D^{t+1}$, the review $C$ and the response $A$ - along with the diverse types of connections between their text elements. Depending on the granularity \textit{g}, text elements can vary from subsentence-level words and phrases to sentences, paragraphs, or sections.
Text elements of granularity \textit{g} of the old and new documents are noted as $x^{t,g}_j \in D^t$ and $x^{t+1,g}_i \in D^{t+1}$.
Comparing two document versions, edit alignment links elements from the new and old versions. For analytical clarity, aligned elements maintain the same granularity in our study, noted as $e_{ij} = e(x^{t+1,g}_i, x^{t,g}_j)$. 
Each edit alignment $e_{ij}$ is associated with an edit label, $EL_{ij}$ = $(g, EA_{ij}, EI_{ij})$, which specifies the granularity, action, and intent of the edit, explaining \textit{how} and \textit{why} the edit is made to a text element of \textit{g} (§\ref{subsec:rev_dim}). 
When a new text element $x^{t+1,g}_i$ is added or an old one $x^{t,g}_j$ is deleted, the corresponding old or new element is null, noted as $e(x^{t+1,g}_i, null)$ or $e(null, x^{t,g}_j)$.

Given that the reviews and responses are typically brief without rich structure, we focus on the sentence level in those documents. Reviews include requests $c_k \in C$ that may prompt edits. An addressed request $c_k$ is linked to relevant edit $e_{ij}$ as $ec(e_{ij}, c_{k})$.  A single request can lead to multiple edits, while self-initiated edits may not connect to any review request. 
Similarly, an author's response includes sentences $a_n \in A$ summarizing realized revisions. Each $a_n$ connects to its respective edit $e_{ij}$ via the relation $ea(e_{ij}, a_n)$.

\begin{figure}[ht]
  \centering
  \includegraphics[width=0.4\textwidth]{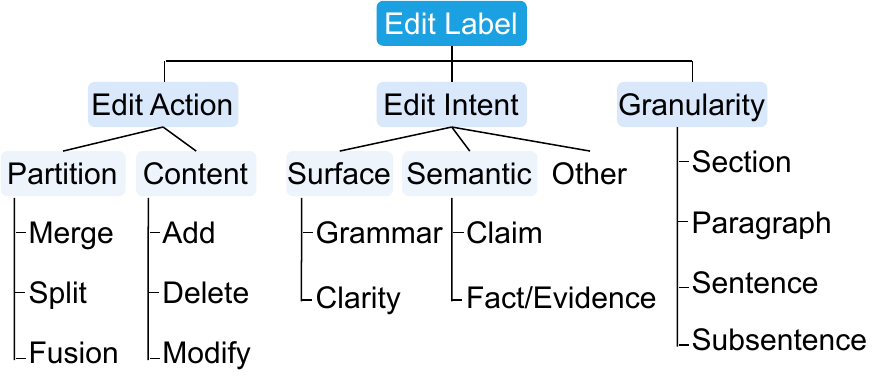}
  \caption[Edit Label]{Re3 label taxonomy. See definitions in §\ref{subsec:rev_dim}.}
  \label{fig:label_taxonomy}
\end{figure}
\subsection{Revision Dimensions and Label Taxonomy} 
\label{subsec:rev_dim}
We analyze revisions along three qualitative dimensions – \textit{granularity}, \textit{action}, and \textit{intent} – and present our proposed label taxonomy in Figure \ref{fig:label_taxonomy}.
\textit{Granularity} specifies the scope of the text elements subject to revision, which is crucial since the perception of revisions varies with granularity. For instance, extending a sentence may appear as adding text elements at subsentence level or as modifying an existing text element at sentence level (further exemplified in Table \ref{tab:3_dim_examples} in §\ref{sec:appendix_label}). 
In this work, we include section, paragraph, sentence, and subsentence granularities.
\textit{Action} specifies how revisions are made, including basic methods like addition, deletion, and modification, as well as complex operations like merges, splits, and fusions (elaborated in Table \ref{tab:edit_actions} in §\ref{sec:appendix_label}). 
\textit{Intent} categorizes the underlying purpose into surface language improvements for grammar or clarity and substantial semantic changes affecting claims or factual content, with detailed definitions and examples in Table \ref{tab:edit_intents} in §\ref{sec:appendix_label}. 
The three dimensions collectively characterize the nature, purpose, and significance of the revisions. For instance, factual content updates may entail sentence expansion with additional details or the incorporation of an entirely new sentence. When significant elaboration is necessary, new paragraphs or sections may be introduced.
The taxonomy has been refined through feedback from two linguists and proved sufficient and crucial in the annotation study with six annotators (§\ref{subsec:anno_process} and §\ref{subsec:appendix_survey}). 
The hierarchical structure of the taxonomy promotes easy expansion and adaptation across various domains by incorporating fine-grained labels.

\section{Dataset Construction: Re3-Sci}
\label{sec:corpus}
\subsection{Data Collection and Pre-processing}
\label{subsec:data_collection}
Scientific publishing, a prominent open source of collaborative document revision and review, offers ample data for our research objectives. 
We instantiate our framework based on the data from the F1000RD dataset \citep{f1000rd} and the ARR-22 subset of the NLPeer corpus \citep{nlpeer}, which include revisions of scientific papers along with their corresponding reviews, covering a range of fields including NLP, science policy, public health, and computational biology. 
Both datasets contain structured documents organized into section and paragraph levels, which we further refine to sentence level (§\ref{subsec:sent_segmentation}). 
A total of 314 document pairs and related reviews are randomly selected for human annotation: 150 from NLPeer and 164 from F1000RD. 

\subsection{Pre-alignment}
\label{subsec:preanno}
Identifying revision pairs from two lengthy documents is challenging, especially complicated by the expansive scope for comparison and the presence of recurring content \cite{jiang_crf}. 
To address this, we employ a lightweight sentence alignment algorithm that systematically excludes identical pairs and identifies alignment candidates from the remaining sentences, considering both form and semantics similarity, as well as the document's context and structure
(§\ref{subsec:sent_pre_alignment}).
Annotators are given the alignments and tasked with validating and correcting any alignment errors. Based on these corrections, the proposed algorithm achieves an accuracy of 0.95. 
The validated alignments are subsequently used for edit action and intent labeling.

\subsection{Annotation Process}
\label{subsec:anno_process}
To perform the human annotation, we develop a cross-document annotation environment using INCEpTION \citep{inception}, as detailed in §\ref{subsec:appendix_inception}.
A pilot study with 20 document pairs is initiated to refine the label taxonomy, optimize the pre-alignment algorithm, improve the annotation tool, and develop comprehensive guidelines, with the assistance of three in-house annotators skilled in computer science or linguistics.

For annotation, six master's students with C1-level English proficiency are recruited (§\ref{subsec:annotators}). 
We employ an iterative data quality management process to ensure the quality of the annotations. Initially, a 15-hour training session is spread over three days, involving a joint review of guidelines, live demonstrations, and practice annotations on a validation set of five document pairs. Given the initial suboptimal IAA in intent labeling, highlighting its complexity, we conduct further discussions on disagreements and common mistakes, followed by a final re-annotation of the validation set. This method ensures consistent comprehension of the guidelines and familiarity with the annotation process prior to actual annotation.
Documents are divided into three data packages for iterative quality assessment, with intermediary meetings by the coordinator to address annotators' individual questions.  
The primary tasks, sentence-level revision alignment and edit labeling, are carried out by three annotators per sample. 
We release all three annotations with a gold label aggregated through majority voting. 
After annotation, we conducted an annotator survey to gather insights for future annotation studies.

We achieve a substantial \citep{Landis1977TheMO} IAA of 0.78 Krippendorff’s $\alpha$ for the labeling task and a perfect IAA of 1 Krippendorff’s $\alpha$ for the alignment task. 
Table \ref{tab:iaa1} in §\ref{subsec:appendix_iaa} shows progressive IAA improvement following iterative quality management between data packages, highlighting the method's effectiveness.
As a qualitative assessment, the annotator survey (§\ref{subsec:appendix_survey}) confirms the adequacy of guidelines, label taxonomy, and annotation tool, as well as the effectiveness of iterative training. The annotators also highlight the effectiveness of the cross-document annotation environment, especially in aligning revision pairs, which potentially contributes to the perfect IAA in alignment. Further insights are provided in §\ref{subsec:appendix_survey}.

\subsection{Statistics}
\label{subsec:anno_results}
\begin{table}[ht]
\fontsize{10}{10}
\selectfont
\centering
\renewcommand{\arraystretch}{1.1}
\tabcolsep=0.15cm
\begin{subtable}[h]{0.47\textwidth}
\begin{tabular}{llllll}
\toprule
  &\#doc &
  \begin{tabular}[c]{@{}l@{}}\# S \end{tabular} &
  \begin{tabular}[c]{@{}l@{}}\# SS \end{tabular} &
  \begin{tabular}[c]{@{}l@{}}\# P  \end{tabular} &
  \begin{tabular}[c]{@{}l@{}}\# Sec \end{tabular} \\ \hline  
  Re3-Sci&314&11,648&2,676&5,064&2,008 \\ \bottomrule
\end{tabular}
\caption{Count of aligned and labeled edits at sentence (S), subsentence (SS), paragraph (P), and section (Sec) levels.}
\label{tab:stat1}
\end{subtable}
\begin{subtable}[h]{0.47\textwidth}
\begin{tabular}{lllll}
\toprule
 &
\begin{tabular}[c]{@{}l@{}}\#review \\request \end{tabular} &
  \begin{tabular}[c]{@{}l@{}}\#review\\-revision \end{tabular} &
  \begin{tabular}[c]{@{}l@{}}\#revision\\-response \end{tabular} \\ \hline 
    Re3-Sci&560&413&1,364 \\ \bottomrule
\end{tabular}
\caption{Count of extracted review requests, their alignments with realized revisions, and linkages between revisions and edit summaries in response.}
\label{tab:stat2}
\end{subtable}
\caption{Re3-Sci dataset statistics.}
\label{tab:stat}
\end{table}
\noindent The Re3-Sci dataset comprises 314 document revision pairs. 11,648 sentence-level edits comprising sentence revision pairs, additions, and deletions are identified and annotated with respective edit action and intent labels. Based on the sentence-level annotations, 5,064 paragraph-level, and 2,008 section-level edits are identified.
We also extract 2,676 subsentence-level edits from 1,453 sentence revision pairs, employing a constituency tree-based method similar to \citet{arxivedits}. 
These extractions and alignments are verified by a linguistic expert and labeled by three annotators. 
Furthermore, we randomly select 42 documents and extract 560 review sentences that may prompt changes. The review sentences are aligned with the corresponding revisions when possible, resulting in 413 linkages. Annotators summarize the document revisions in brief responses and then align a total of 784 summary sentences back to the related edits, resulting in 1,364 connections. See §\ref{subsec:annotators} for more details.

\begin{figure}[ht]
\centering
\includegraphics[width=0.37\textwidth]{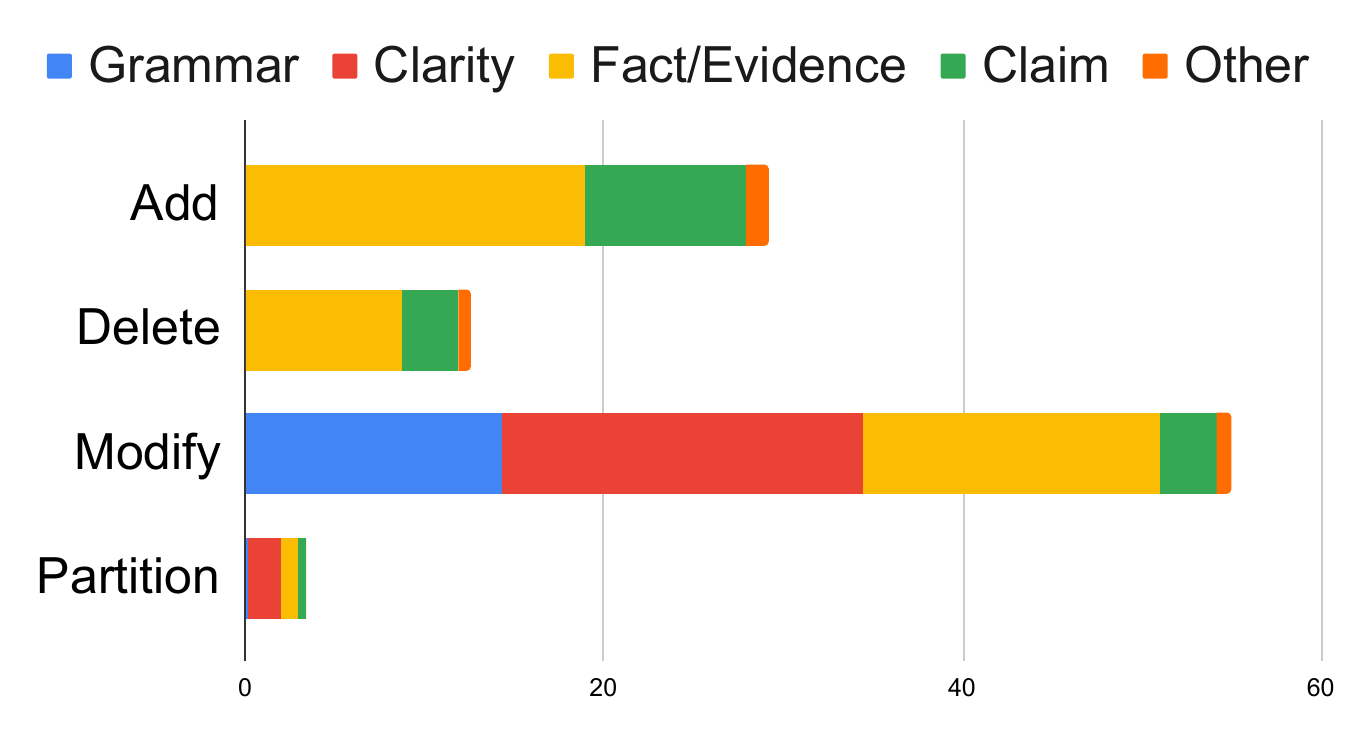}
  \caption{Proportions of sentence edit action and intent labels and their combinations in Re3-Sci.}
  \label{fig:ea_ei_comb}
\end{figure}

\section{Dataset Analysis}
\label{sec:analysis}
The framework and dataset allow us to answer new questions about human behavior in collaborative document revision in scholarly publishing.

\noindent\textbf{RQ1: How and why do humans edit, and what are the relationships between edit actions and intents?} 
Figure \ref{fig:ea_ei_comb} reveals that authors predominantly modify existing content and add new material, with deletions being infrequent and partition changes even less common. It also suggests that the enhancement of fact or evidence is the primary focus of revisions, highlighting its importance in improving scientific quality. 
Moreover, Figure \ref{fig:ea_ei_comb} illustrates that additions and deletions of sentences typically pertain to improving factual content or claims, but are never intended for superficial language enhancement. On the other hand, grammar and clarity improvements are usually realized by modifying existing sentences. 
This suggests that, from a modeling view, the edit action and intent labels may influence the prediction of each other.

\begin{figure}[ht]
  \centering
  \begin{subfigure}[b]{0.49\columnwidth}
    \includegraphics[width=\linewidth]{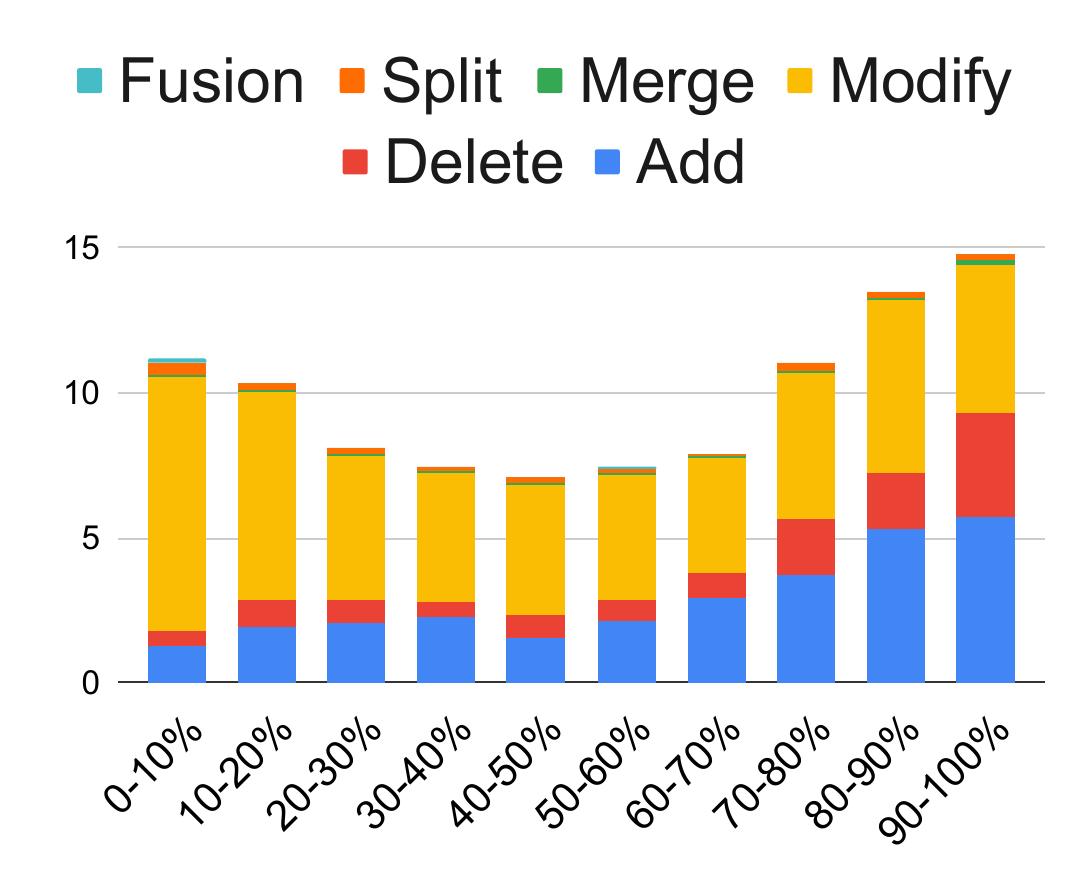}
    \caption{Edit action distribution}
    \label{fig:ea_over_doc}
  \end{subfigure}
  \hfill
  \begin{subfigure}[b]{0.49\columnwidth}
    \includegraphics[width=\linewidth]{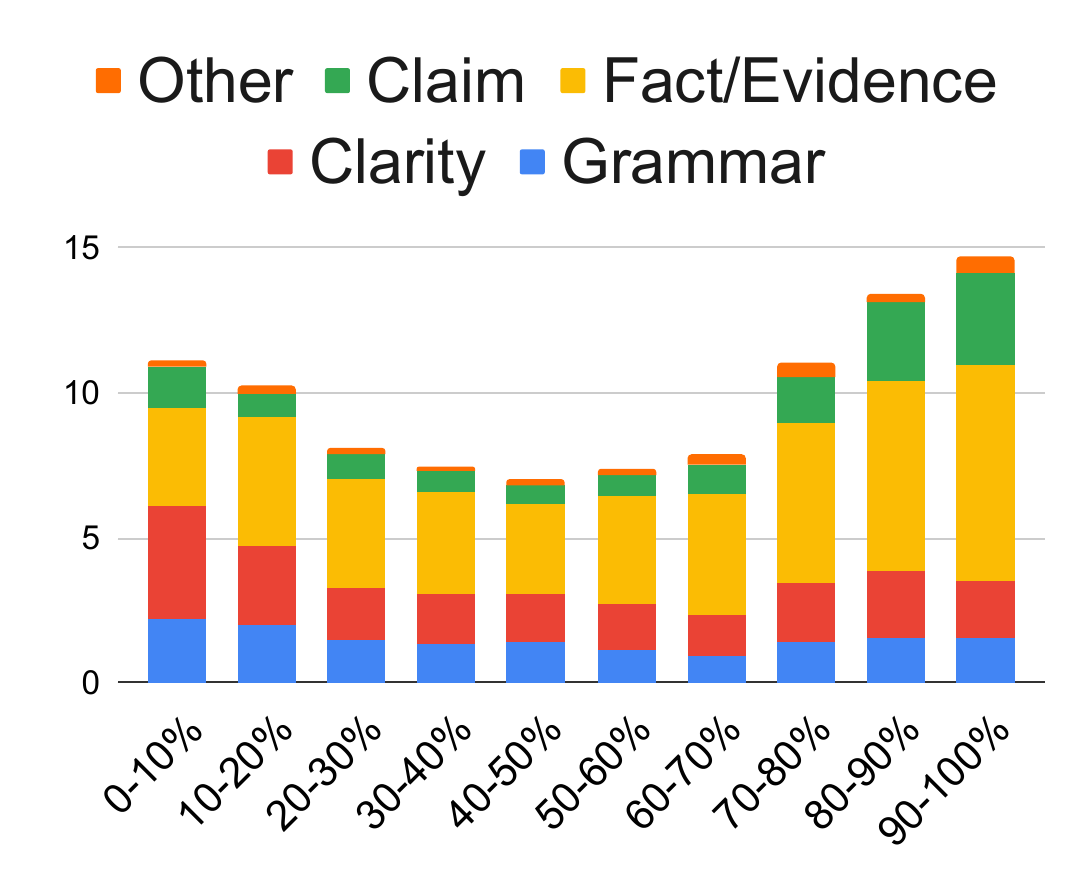}
    \caption{Edit intent distribution}
    \label{fig:ei_over_doc}
  \end{subfigure}
  \caption{Edit action and intent distribution over the document. The x-axis represents the relative sentence positions within documents.}
  \label{fig:edits_over_doc}
\end{figure}
\noindent\textbf{RQ2: How are edits distributed in documents?} 
Figure \ref{fig:edits_over_doc} indicates that the initial and final parts of papers experience significantly more revisions. In terms of edit actions, the beginning of the document typically sees more modifications, while the end is characterized by a higher frequency of additions and deletions. Regarding edit intents, language enhancements for grammar or clarity are more common in the early parts, whereas changes affecting semantic content, such as facts or claims tend to occur more in the later parts. 
These suggest that the document position may be a valuable predictor for identifying edit actions and intents. 

\begin{figure}[ht]
  \centering
  \begin{subfigure}[b]{0.49\columnwidth}\includegraphics[width=\linewidth]{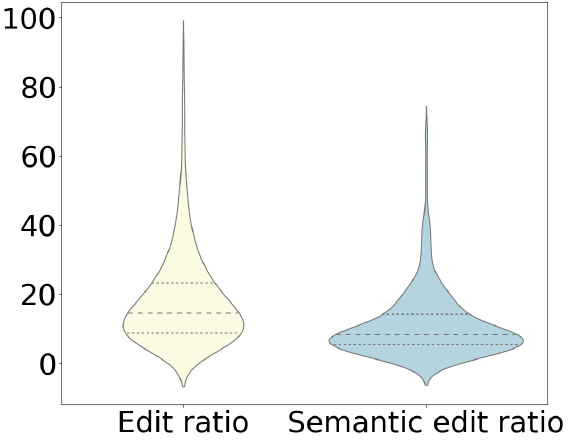}
    \caption{Document Update}
    \label{fig:vio_ratio}
  \end{subfigure}
  \begin{subfigure}[b]{0.48\columnwidth}
    \includegraphics[width=\linewidth]{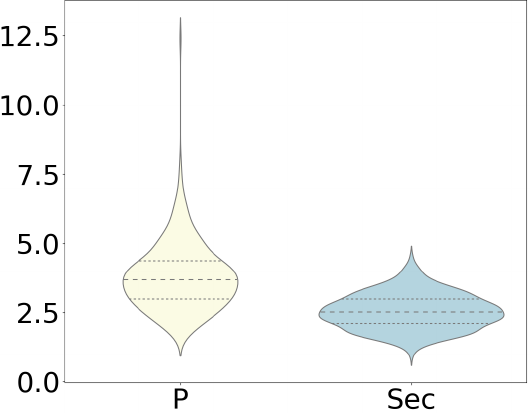}
    \caption{Crest Factor (CF)}
    \label{fig:vio_cf}
  \end{subfigure}
  \caption{(a) Document update measured by edit ratio and semantic edit ratio. (b) Crest Factor (CF) measured at paragraph (P) and section (Sec) level.}
  \label{fig:vio}
\end{figure}

\noindent\textbf{RQ3: How significant are the differences between document versions?} 
To gauge the magnitude of change, we introduce the \textit{Edit Ratio} metric, determined by the ratio of sentence edits to the sentence count in the original document. 
While the edit ratio reflects the extent of differences, the significance of document revisions is highlighted by the \textit{Semantic Edit Ratio}, which is calculated by the ratio of semantic edits labeled with Fact/Evidence or Claim.
The average document edit ratio stands at 18.45\%. Figure \ref{fig:vio_ratio} and Figure \ref{fig:edit_ratio_2} in §\ref{sec:appendix_c} show that the majority of documents experience moderate revisions with an edit ratio of 5-25\%, while a small proportion has an edit ratio exceeding 50\%, and only a few documents appear to have been extensively rewritten.
The average semantic edit ratio stands at 11.18\%, with most documents showing 0-20\% of their content undergoing significant change. Notably, documents with a high edit ratio often do not correspond to a high semantic edit ratio, suggesting that documents with extensive revisions typically exhibit language quality issues.

\noindent\textbf{RQ4: How are edits clustered by paragraphs and sections?}
We use Crest Factor (CF), a concept borrowed from signal processing \citep{parker2017digital}, to assess the concentration of edits.
Using a vector of sentence edit counts in each paragraph or section, CF quantifies the peak amplitude of this distribution. A CF value of 1 signifies an even distribution.
The average paragraph CF is 3.79, indicating a substantial concentration of edits within a limited number of paragraphs. 
This trend of high edit concentration in a few paragraphs is further illustrated in Figure \ref{fig:vio_cf} and Figure \ref{fig:cf} in §\ref{sec:appendix_c}.
When examined at the section level, the average CF is 2.54, indicating a moderate tendency towards clustering.

\begin{figure}[ht]
\centering
\includegraphics[width=0.37\textwidth]{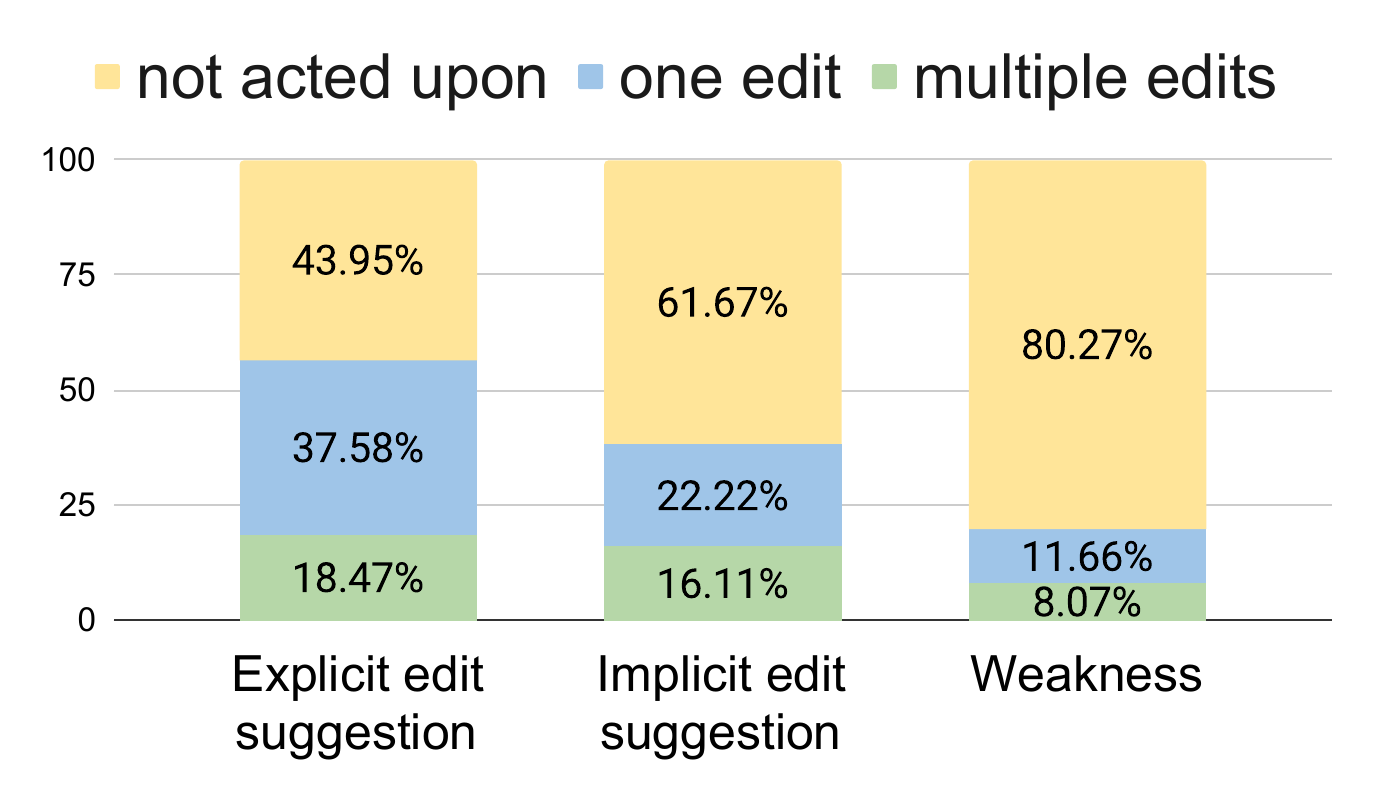}
  \caption{Three types of reviewers' requests and their impact. Displayed are proportions of requests not acted upon,  actualized by single or multiple sentence edits.}
  \label{fig:sug_taken}
\end{figure}

\noindent\textbf{RQ5: Are reviewers' requests acted upon? How are these realized in revision?} 
Annotators categorize the relevant review requests into three types: \textit{explicit edit suggestions} (28\%),  
 \textit{implicit edit suggestions} (32.1\%), and \textit{general weakness comments} (39.8\%). 
The first provides specific document locations and clear revision instructions; the second delivers guidance without locations; the last highlights general issues without specific suggestions.
Figure \ref{fig:sug_taken} shows that more than half of explicit suggestions are implemented, with 18.47\% actualized through multiple sentence edits. Implicit suggestions and general weakness comments are realized to a lesser extent. This implies that reviewers' explicit suggestions are more likely to be acted upon.

\section{Automation with LLMs}
\label{sec:models}
The Re3-Sci dataset facilitates a variety of NLP tasks, such as (1) \textit{edit intent classification}, (2) \textit{document edit summarization}, (3) \textit{revision alignment} and (4) \textit{review request extraction}. 
Tasks (1) and (3) demonstrate the joint modeling of old and new documents (revision), with the latter providing the basis for NLP-assisted edit analysis and the former enabling in-depth analysis and being challenging even for human annotators (§\ref{subsec:appendix_survey}).
Task (2) represents a novel task to jointly model the revision-response process, a capability uniquely enabled by our new dataset. Re3-Sci is distinguished by its full-scope annotations, which cover all edits with action and intent labels, and include information on the document's context, structure, and granularity (§\ref{sec:corpus}). These features enable detailed descriptions of edits and precise localization of their positions within the document (§\ref{subsec:task_summ}).
Task (4) focuses on the review-revision relationship, aiming to identify review sentences that need attention during the revision phase.
The dataset's utility extends beyond these four tasks,  including edit generation from review requests or desired actions/intents, and anchoring edits to review requests, which we leave for future work.
§\ref{subsec:appendix_task_gpu} provides computational details.

\subsection{Edit Intent Classification}
\label{subsec:task_intent}
\noindent\textbf{Task formulation and data.} 
Formulated as a classification task given a sentence edit $e(x^{t+1,g}_i, x^{t,g}_j)$, the objective is to predict the intent label $EI_{ij}$.
For additions or deletions, only one sentence is used. 
We split the documents into 20\% for training and 80\% for testing.\footnote{We use training data for ICL example selection only and the rest for testing to get more reliable performance estimates.} The test set contains 5,045 revision pairs and 3,891 additions or deletions.

\noindent\textbf{Models and methods.} 
We evaluate Llama2-70B \cite{llama2} with multiple ICL demonstration selection methods and analyze CoT prompt formatting.
The three dynamic ICL methods select the most similar demonstrations from the training set for each test sample using RoBERTa embeddings \cite{roberta}: \textit{cat} uses cosine similarity of concatenated sentence embeddings, \textit{diff} leverages the difference between sentence embeddings, and \textit{loc} utilizes concatenated embeddings of the associated section titles.
The static \textit{def} method uses a default set of manually selected examples for each intent across all tests.
For CoT, we instruct the model to predict the intent label (\textit{L}) with rationale (\textit{R}) in CoT style, evaluating how their order impacts results. A prompt example is provided in Table \ref{tab:prompt} in §\ref{subsec:appendix_task_intent}.
Preliminary experiments indicate inadequate performance in jointly modeling revision pairs and single-sentence instances, leading us to separate experiments for each scenario.

\noindent\textbf{Results and discussion.} Table \ref{tab:m_intent} shows the results on revision pairs.
In addition to the random baseline, other baselines use the majority label of the top \textit{n} selected training examples from the three proposed methods.
Using the same examples for ICL, the \textit{diff} method notably excels over others (block 1).
Interestingly, Llama2 doesn't rely solely on the majority label of selected examples. Comparing block 1 with the majority baselines reveals a significant improvement and reduced disparities between the methods. 
Using five default examples outperforms \textit{cat} and \textit{loc}, and is on par with \textit{diff} (block 2). Accuracy further increases when the gold label is accompanied by a rationale in CoT style (i.e., L,R: label followed by rationale). This straightforward but effective prompting method achieves performance comparable to more advanced methods, as detailed in the subsequent blocks of Table \ref{tab:m_intent}.
However, reversing the order of the label and the rationale (i.e., R,L) notably decreases performance.
Combining default examples with rationale and dynamic \textit{diff}-selected examples further enhances accuracy (block 3). Altering the order of dynamic and static default examples enhances results when using \textit{diff}, though this is not consistent across all selection methods (blocks 3, 4). 
Omitting rationale from the default examples leads to a significant and consistent performance decline, highlighting the importance of CoT demonstrations (blocks 5, 6).

\begin{table}[t]
\fontsize{8}{8}
\selectfont
\centering
\renewcommand{\arraystretch}{1.16} 
\begin{tabular}[]{ll|ll|ll}
\hline
\multicolumn{6}{c}{\textbf{Baselines}} \\ \hline
Random   & 0.20&&& \\
diff1  & 0.45  & cat1 & 0.38  & loc1 & 0.31 \\
diff3-maj & 0.45  & cat3-maj & 0.38  & loc3-maj & 0.32 \\
diff5-maj  & 0.46  & cat5-maj & 0.40  & loc5-maj & 0.34 \\
diff8-maj  & \underline{0.47}  & cat8-maj & 0.41  & loc8-maj & 0.33 \\
\hline
\multicolumn{6}{c}{\textbf{Our Models (ICL \& CoT)}} \\ \hline
\multicolumn{6}{l}{\textcircled{\tiny 1}\textit{+ dynamic examples}}  \\ 
+diff1  & 0.60  & +cat1 & 0.58  & +loc1 & 0.56 \\
+diff3 & 0.60  & +cat3  & 0.57  &+loc3  & 0.53 \\
+diff5  & \underline{0.61} & +cat5& 0.56  & +loc5 & 0.52  \\
+diff8  & 0.59  & +cat8 & 0.56   & +loc8 & 0.51 \\ \hline
\multicolumn{6}{l}{\textcircled{\tiny 2}\textit{+ static examples}}  \\ 
+def5  & 0.59  & \begin{tabular}{l}
     +def5\\-(L,R)
\end{tabular} & \underline{0.62}  & \begin{tabular}{l}
     +def5\\-(R,L)
\end{tabular}   & 0.53 \\ \hline
\multicolumn{4}{l}{\textcircled{\tiny 3}\textit{+ def5-(L,R) + dynamic }}   \\ 
+diff1 & 0.62 &+cat1 & 0.59 &+loc1 &0.59 \\
+diff3 & \underline{0.63} &+cat3 & 0.59 &+loc3 &0.58 \\
+diff5 & \underline{0.63} & &  & & \\ \hline
\multicolumn{4}{l}{\textcircled{\tiny 4}\textit{+ dynamic + def5-(L,R)}} \\ 
+diff1 & 0.64 &+cat1 & 0.58 &+loc1 &0.60 \\
+diff3 & \underline{\textbf{0.65}} &+cat3 & 0.59 &+loc3 &0.58 \\
+diff5 & 0.63 & &  & & \\ \hline
\multicolumn{4}{l}{\textcircled{\tiny 5}\textit{+ def5 + dynamic }}  \\ 
+diff1 & 0.59 &+cat1 & 0.58 &+loc1 &0.57 \\
+diff3 & \underline{0.61} &+cat3 & 0.57 &+loc3 &0.55 \\
+diff5 & 0.59 & &  & & \\ \hline
\multicolumn{4}{l}{\textcircled{\tiny 6}\textit{+ dynamic + def5 }} \\ 
+diff1 & 0.59 &+cat1 & 0.57 &+loc1 &0.55 \\
+diff3 & 0.61 &+cat3 & 0.56 &+loc3 &0.54 \\
+diff5 & \underline{0.62} & &  & & \\ \hline
\end{tabular}
\caption[LLMs]{Llama2-70B accuracy in edit intents classification on revision pairs. 
Baselines are assessed on the full test set, subsequent models are evaluated on 20\% of test samples for validation. 
Underlined is the best accuracy in the block, the highest accuracy is in bold.
}
\label{tab:m_intent}
\end{table}

\begin{figure}[ht]
  \centering
  \begin{subfigure}[b]{0.48\columnwidth}
    \includegraphics[width=\linewidth]{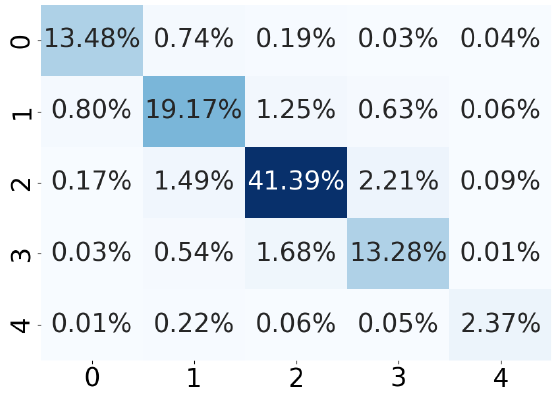}
    \caption{Human}
    \label{fig:intent_human}
  \end{subfigure}
  \hfill
  \begin{subfigure}[b]{0.48\columnwidth}
    \includegraphics[width=\linewidth]{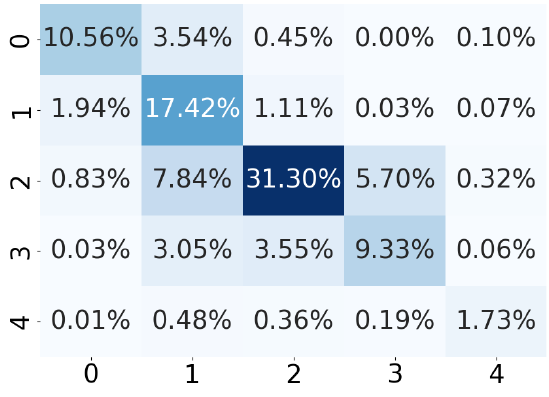}
    \caption{Llama2-70B}
    \label{fig:intent_gpt}
  \end{subfigure}
  \caption{Error analysis and human vs. LLM comparison in edit intent classification on the full test set. The y-axis presents the gold, the x-axis presents the annotated/predicted labels. The gold labels are the majority of the three human labels. The diagonal indicates the percentages of correct labels. 0: Grammar, 1: Clarity, 2: Fact/Evidence, 3: Claim, 4: Other.}
  \label{fig:intent_compare}
\end{figure}

\begin{table*}[ht]
\fontsize{8}{8}
\tabcolsep=0.08cm
\renewcommand{\arraystretch}{1.1} 
\begin{tabular}[t]{l|ll|llll|l }
\hline
 & \#S & \#W 
&Factuality & Comprehensiveness & Specificity & Compactness & Organization \\\hline
human  &19 & 346  &100\% &98.82\%  &95.56\% & 1.74 &100\% section\\ \hline
GPT-4  & 16 &309 & 95.96\% &79.09\% &89.82\% &2.36
& \begin{tabular}{l}72.5\% action, 17.5\% section
\end{tabular}\\ \hline
\end{tabular}
\caption[LLMs]{Human evaluation and human vs. LLM comparison in document edit summarization. 
Demonstrated are the average counts of summary sentences (\#S) and words (\#W), as well as the five measures (§\ref{subsec:task_summ}).}
\label{tab:edit_summ}
\end{table*}

The best configuration involves three dynamic \textit{diff}-selected examples and the default examples with CoT rationale.
This also yields the best performance for additions and deletions, as shown in Table \ref{tab:ad_intent} in §\ref{subsec:appendix_task_intent}.
With this setup, joint evaluation on the full test set results in an accuracy of 0.7 and a macro-average F1 score of 0.69, significantly outperforming the baselines as shown in Table \ref{tab:adm_intent} in §\ref{subsec:appendix_task_intent}. 
The pronounced potential for advancement highlights the task's complexity for LLMs. This requires precise detection of changes and advanced reasoning capabilities to understand intents.

Figure \ref{fig:intent_compare} displays an error analysis comparing human annotations with Llama2 predictions. Both humans and Llama2 are prone to misclassify claim and fact changes, which may stem from subjective statements being phrased in a fact-like manner and the common occurrence of intertwining both aspects within a single sentence. Llama2 demonstrates a propensity to over-predict clarity changes, often misinterpreting fact/evidence (7.84\%), grammar (3.54\%), and claim (3.05\%) changes.

\subsection{Document Edit Summarization}
\label{subsec:task_summ}
\noindent\textbf{Task formulation and data.}  
Our full-scope annotations of document revisions enable a novel task, document edit summarization, which constitutes the foundational basis for generating author responses.  
Specifically, the task is formulated as a text generation task, given a complete list of sentence edits $e(x^{t+1,g}_i, x^{t,g}_j)$ within a document, associated action and intent labels $EL_{i,j}$, as well as associated section titles that provide structural information about the edits.
The output is a coherent textual summary of the document edits, as exemplified in Table \ref{tab:edit_summ_example} in §\ref{subsec:appendix_edit_summ}.
We conduct experiments on 42 documents with human-written edit summaries. These documents contain an average of 33 sentence edits, resulting in a median input length of 3,916 tokens.

\noindent\textbf{Models and methods.} 
As almost half of the inputs exceed Llama2's constraints and preliminary trials yield unsatisfactory results, we opt for GPT-4 to handle this challenging task in a zero-shot manner.
We performed a human evaluation of the generated summaries, systematically comparing them with human-authored summaries across five dimensions: \textit{factuality}, \textit{comprehensiveness}, \textit{specificity}, \textit{compactness}, and \textit{organization}. 
Each summary sentence $a_n$ is linked to its respective sentence edits, creating $ea$ linkages. And $a_n$ is annotated if it does not refer to actual edits and labels. or it is hard to connect with specific edits. For instance, vague summaries such as "there are grammar corrections" pose challenges in establishing precise associations. 
\textit{Factuality} is quantified by the percentage of summary sentences that accurately refer to actual edits. \textit{Comprehensiveness} denotes the extent of edits encapsulated within the summary. \textit{Specificity} reflects the proportion of concrete summary sentences. \textit{Compactness} is gauged by the average number of edits incorporated into a single summary sentence. And \textit{organization} refers to the logical arrangement of the summary content. 

\noindent\textbf{Results and discussion.} 
Table \ref{tab:edit_summ} provides a comparative analysis between human-authored and LLM-generated summaries. Humans typically produce marginally lengthier text with more sentences, ensuring impeccable factuality alongside elevated comprehensiveness and specificity. Conversely, GPT-4 fails to address 21\% of document edits, exhibiting factuality concerns in 4\% of summary sentences, and lacking specificity in 10\% of cases. Additionally, GPT-4 summaries demonstrate a slightly higher level of compactness, averaging 2.36 edits condensed into a single sentence.
While humans typically organize summaries by sections, reflecting conventional sequential reading patterns. GPT-4 also exhibits a structured logical arrangement but often organizes summaries by action labels, usually beginning with additions and deletions.

\subsection{Revision Alignment}
\label{subsec:appendix_task_alignment}
\noindent\textbf{Task formulation and data.} 
The task is conceptualized as a binary classification problem, where the goal is to determine if a given pair of sentences  $x^{t+1,g}_i \in D^{t+1}$, $x^{t,g}_j \in D^{t}$ constitutes a revision pair $e(x^{t+1,g}_i, x^{t,g}_j)$. 
Along with the 6,353 revision pairs from the Re3-Sci dataset, an equivalent number of negative samples are created, resulting in a total of 12,706 samples for experimental purposes. 80\% are used for testing and 20\% for training. 
To preserve the task's complexity, negative samples are composed by pairing revised sentences within the same document that do not link to each other but likely address similar topics.  This simulates the intricate nature of revisions in lengthy documents as detailed in §\ref{subsec:sent_pre_alignment}.

\noindent\textbf{Models and methods.}
For this task, we employ the Llama2-70B model and apply the same ICL and CoT methods used for the edit intent classification task, as detailed in §\ref{subsec:task_intent}.

\noindent\textbf{Results and discussion.} 
Mirroring the same findings observed in the edit intent classification tasks, Table \ref{tab:revision_align} in §\ref{subsec:appendix_align} shows that using static default examples with CoT reasoning throughout the experiments yields favorable performance (block 2), highlighting its efficacy as a straightforward yet effective prompting strategy. Using this strategy, we achieve an accuracy of 0.97 on the full test set.

It is worth noting that our proposed pre-alignment algorithm (§\ref{subsec:sent_pre_alignment}) achieves a strong accuracy of 0.95, with a recall of 0.99 for non-alignment and a precision of 0.99 for alignment. However, the precision for non-alignment (0.89) and the recall for alignment (0.92) are relatively low. This discrepancy can be attributed to the utilization of high similarity thresholds and stringent aligning rules in the algorithm. In contrast, when automated with Llama2, we achieve a precision of 0.99 for non-alignment and a recall of 0.99 for alignment, which constitutes a perfect enhancement to the pre-alignment algorithm. For revision alignment, we thus propose \textbf{a two-stage approach} that combines the lightweight pre-alignment algorithm with Llama2 In-Context learning. The lightweight algorithm efficiently identifies candidates and accurately extracts revision pairs with minimal computational cost. Subsequently, we apply the proposed prompting strategy with Llama2 selectively to the non-aligned candidates, thereby identifying missing revision pairs without significantly increasing computational overhead.

\subsection{Review Request Extraction}
\label{subsec:appendix_task_review} 
\noindent\textbf{Task formulation and data.} The task is framed as a binary classification problem, aiming to ascertain whether a particular review sentence $c_k \in C$ could instigate revisions and necessitate further processing in the revision workflow.
The experimental data comprises 1,000 samples, including 560 review requests (including explicit and implicit edit suggestions, and general weakness comments) from the Re3-Sci dataset, plus 440 negative samples extracted from the same review documents. Of these, 80\% are for testing and 20\% for training.

\noindent\textbf{Models and methods.} For this task, we utilize Llama2-70B with the same ICL and CoT methods previously applied, as elaborated in §\ref{subsec:task_intent}.

\noindent\textbf{Results and discussion.} 
Employing the straightforward \textit{def} method with CoT reasoning, which involves two static default demonstrations, yields an accuracy of 0.80 on the full test set. This approach achieves a high precision of 0.95 for negative samples and a remarkable recall of 0.98 for positive samples. Nevertheless, the precision for positive samples is relatively low at 0.74, highlighting the method's inherent challenges.
Future research could expand this task into a four-label classification, differentiating various types of review requests. This approach could further elucidate the methods' capabilities and limitations.

\section{Conclusion}

\noindent We have introduced the Re3 framework and the Re3-Sci dataset, for empirical analysis and development of NLP assistance for text-based collaboration.
Through annotation study and data analysis, we have demonstrated the utility of the framework and revealed novel insights into human behavior in collaborative document revision and peer review, including relationships between specific edit actions and intents, focused localization patterns, clustering tendencies within paragraphs, and the acceptance rates for review requests. 
Our automation experiments have assessed the ICL and CoT capabilities of state-of-the-art LLMs on four tasks for collaborative revision assistance.  
In the classification tasks with Llama2-70B, we noted that using default static ICL demonstrations with CoT rationale produces satisfactory results, demonstrating the efficacy of this straightforward yet effective prompting strategy. 
In the document edit summarization task, GPT-4 demonstrated the ability to generate coherent summaries but faced challenges related to factuality and comprehensiveness.

Our work paves the path towards systematic full-scope study of text-based collaboration in NLP and beyond. The framework, taxonomy, annotation methods and tools are applicable to diverse domains. The dataset offers a robust foundation for multifaceted research for collaborative revision assistance. Future work may encompass tasks like identifying text segments necessitating revision and generating revisions guided by review requests or specified actions and intents.

\section*{Limitations}
This study has several limitations that should be considered when interpreting our results and the implications we draw from them. From the data and modeling perspective, the study's exclusive focus on English-language scientific publications is due to the restricted availability of openly licensed source data. Studying the transferability of our findings to new languages, domains, application settings and editorial workflows is an exciting avenue for future research, which can be supported by our openly available annotation environment and protocols.
Our study used human-generated edit summaries instead of author responses or summaries of changes written by the authors themselves due to the lack of data. As peer-reviewing data collection becomes increasingly popular in the NLP community, we expect new datasets to enable such studies in the future.

From a task perspective, it is important to highlight that the implementations and results presented in this study serve as illustrations of the proposed tasks. Their primary purpose is to ascertain the technical feasibility and lay the groundwork for the development of future NLP systems for collaborative writing and revision assistance. Consequently, the provided implementations have inherent limitations. For instance, our approach selectively utilizes state-of-the-art LLMs without conducting comprehensive comparisons with other LLMs or smaller fine-tuning-based models. A systematic exploration of NLP approaches for the proposed tasks lies beyond our scope and is left for the future.

\section*{Ethics Statement} 
The analysis of text-based collaboration and the corresponding NLP assistance applications have a potential to make knowledge work more efficient across many areas of human activity. We believe that the applications and analysis proposed in this paper deliver equitable benefits to every stakeholder involved in the procedure -- authors, co-authors, reviewers, and researchers who want to study their collaborative text work.

The human annotators employed in our study were fairly compensated with a standard salary for student assistants in the country of residence. They were informed and consented to the publication of their annotations as part of this study. The annotation process does not entail the gathering or handling of their personal or sensitive information.
For privacy protection, both author metadata and annotator identities have been omitted from the data release.

Both subsets of the source data are licensed under CC-BY-NC 4.0, ensuring that the construction and use of our dataset comply with licensing terms. Our annotated Re3-Sci dataset is available under a CC-BY-NC 4.0 license.

\section*{Acknowledgements}
This study is part of the InterText initiative\footnote{\url{https://intertext.ukp-lab.de/}} at the UKP Lab. This work has been funded by the German Research Foundation (DFG) as part of the PEER project (grant GU 798/28-1) and co-funded by the European Union (ERC, InterText, 101054961). Views and opinions expressed are however those of the author(s) only and do not necessarily reflect those of the European Union or the European Research Council. Neither the European Union nor the granting authority can be held responsible for them.

We extend our sincere gratitude to Dr.-Ing. Richard Eckart de Castilho for his invaluable assistance in creating the cross-document annotation environment within INCEpTION. We also express appreciation to our research assistants and annotators: Gabriel Thiem, Sooyeong  Kim, Xingyu Ma, Manisha Thapaliya, Valentina Prishchepova, Malihe Mousarrezaei Kaffash, and ABM Rafid Anwar for their dedicated efforts, active involvement, and valuable feedback throughout the entire process.


\appendix

\section{Label Taxonomy and Examples}
\label{sec:appendix_label}
Table \ref{tab:3_dim_examples} presents examples of revisions analyzed according to the three dimensions: granularity, action, and intent, illustrating their importance and indispensability. Table \ref{tab:edit_actions} and Table \ref{tab:edit_intents} offer detailed definitions and examples for the edit action labels and the edit intent labels, respectively.

\section{Annotation}
\label{sec:appendix_b}
\subsection{Sentence Segmentation}
\label{subsec:sent_segmentation}
Both F1000RD and NLPeer datasets contain structured documents as intertextual graphs (ITG), a comprehensive document representation format that maintains document structure, cross-document links, and granularity details \citep{f1000rd}. 
In those ITGs, paragraphs are the most refined text elements.
For our study, we opt to commence with more granular units of sentences. This creates a solid baseline for subsequent expansion to broader units, or to microscopic subdivisions.

\begin{algorithm}[ht]
\fontsize{11}{11}
\selectfont
\renewcommand{\arraystretch}{1.4} 
\tabcolsep=0.1cm
\textbf{Input} : $x_i^{t+1,g} \in D^{t+1}$, $x_j^{t,g} \in D^t$, g = S\\
\textbf{Output} : 
$alignS \in 1^{k \times  l}$ to $0^{k \times  l}$ \\
\textbf{Ensure} : $0 < t0,t1 < 100$,  $t0 < t1$
\begin{algorithmic}
\For{$i \gets 1$ to $k$} 
\For{$j \gets 1$ to $l$} 
\For{$m \in M$}
\State $simS[m,i,j] \gets m(x_i^{t+1,g},x_j^{t,g})$
\EndFor
\EndFor
\EndFor
\For{$i \gets 1$ to $k$}  
\For{$m \in M$}
\State \[j_{max} = \argmax_{i,m} simS[m,i,j]\]
\If {$simS[m,i,j_{max}] > t1$ \\   and $all(simS[m,i,j_{max}] > t0, m\in M)$}  
\State $C_i \gets C_i+j_{max}$ 
\EndIf
\EndFor
\If {$len(f(C_i))==1$} 
\State $j_{align} = f(C_i)[0]$
\State $alignS[i, j_{align}]=1$
\ElsIf {$len(f(C_i))>1$} 
\State \[j_{align} = \argmin_i d(i,j), j \in f(C_i)\]
\State $alignS[i, j_{align}]=1$
\EndIf
\EndFor
\end{algorithmic}
\caption{Sentence pre-alignment algorithm}\label{alg:preanno}
\end{algorithm}

We augment the original ITG documents with sentence nodes, employing an assembled sentence segmentation methodology using spaCy\footnote{Version 3.2.4} 
and ScispaCy\footnote{We use the implementation provided at:  \url{https://github.com/allenai/scispacy/blob/main/scispacy/custom_sentence_segmenter.py}} \citep{scispacy}. In our preliminary testing, we discovered that neither spaCy nor ScispaCy sentence splitters are infallible for segmentation, with neither consistently outperforming the other. They can erroneously segment text based on punctuation, such as dots, which are critical for accurate revision alignment. For instance, a dot within numerical values in a sentence could trigger an incorrect segmentation and result in two sentence units. If this dot is omitted in the new version, the sentence is correctly extracted, leading to significant challenges and errors in aligning the two sentences as a revision pair. We employ an assembly of the two sentence splitters, opting for fewer segmentations yielding a smaller number of longer sentences, which mitigates most incorrect splits. Additionally, special nodes such as article titles, section titles, and list elements are not split. 

\begin{figure*}[ht]
  \centering
  \includegraphics[width=1\textwidth]{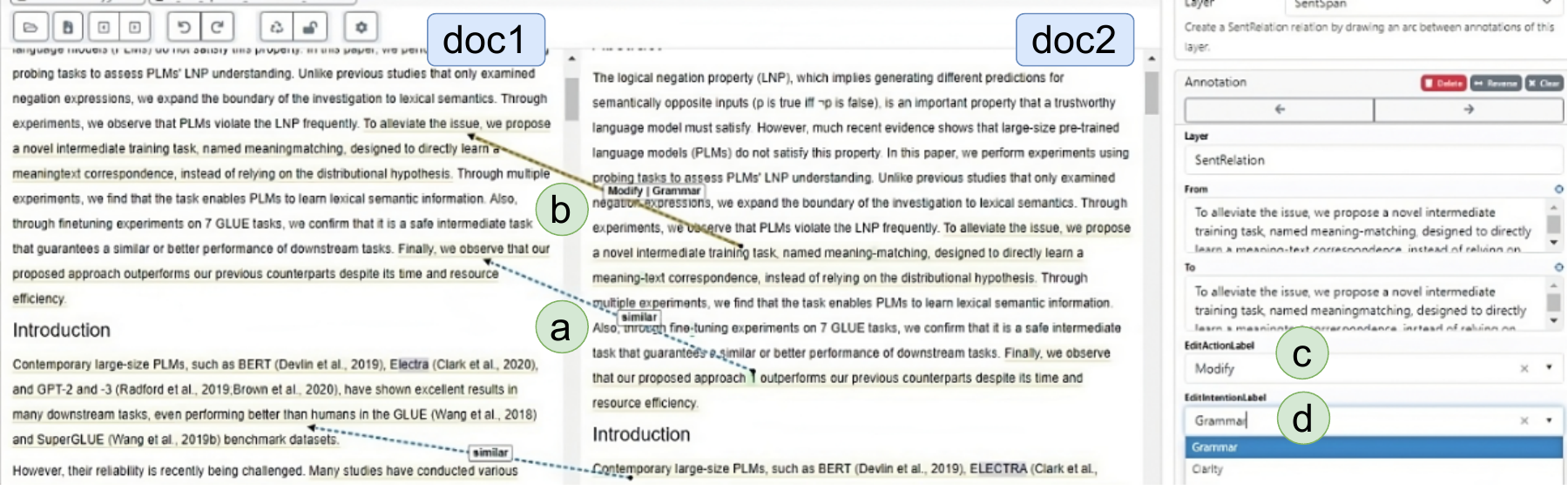}
\caption{INCEpTION enables cross-document annotation in the context of two full documents (doc1-doc2). Annotators can scroll up and down to read the entire document context. The document's structure, including its sections and paragraphs, is preserved. The results from the pre-alignment algorithm (§\ref{subsec:sent_pre_alignment}) are provided (a). Annotators are tasked to validate the pre-alignments (b), and select the edit action (c) and intent (d) labels.}
\label{fig:inception}
\end{figure*}

The segmentations are verified and corrected by a linguistics expert, demonstrating that this integrated approach significantly enhances accuracy compared to using either splitter individually.
\subsection{Sentence Revision Pre-alignment}
\label{subsec:sent_pre_alignment}
Identifying revision pairs from two lengthy documents is challenging, especially complicated by the expansive scope for comparison and the presence of recurring content. 
In lengthy documents, it's crucial to align similar sentences, but it's even more vital to avoid aligning non-relevant sentences with overlapping content. 
To address this, we design a lightweight algorithm to automatically pre-annotate sentence revision pairs, additions and deletions, which achieves a decent accuracy of 0.95. 
The algorithm is detailed in Algorithm \ref{alg:preanno} which follows these steps: 
\begin{enumerate}[itemsep=0pt, parsep=0pt]
\item For sentence-level alignment ($g=S$), after removing identical paragraph pairs, followed by the removal of identical sentence pairs from the remaining text, there remain \textit{k} sentences in the new document $D^{t+1}$ and \textit{l} sentences in the old document $D^{t}$. 
\item For each potential pair, a set of similarity measures $m \in M$ is computed, including Levenshtein distance \cite{Levenshtein1965BinaryCC} and fuzzy string matching\footnote{We use fuzzywuzzy 0.18.0 at: \url{https://github.com/seatgeek/fuzzywuzzy}}, as well as semantic similarity measured by SBERT \citep{sbert}.
\item For each remaining sentence $x_i^{t+1,g} \in D^{t+1}$, using each measure \textit{m}, the algorithm identifies the most similar candidate $x_j^{t,g}$. If the similarity score exceeds threshold \textit{t1} and all other similarity scores between $x_i^{t+1,g}$ and $x_j^{t,g}$ surpass \textit{t0}, $x_j^{t,g}$ is included in the candidate list $C_i$, ensuring pairs similar in \textbf{both} form and meaning are found. 
 \item The function \textit{f} determines the most frequent element in the resulting candidate list $C_i$.  In cases of multiple equally frequent elements due to repeated content, the alignment is assigned to the candidate closest in location,  determined by 
 \begin{equation}
 d_{i,j}= \mid \frac{p_i}{\#P^{t+1}}- \frac{p_j}{\#P^{t}}\mid
 \end{equation}
 where $p_i$  is the linear index of the paragraph containing $x_i^{t+1,g}$, and $\#P^{t+1}$ is the total number of paragraphs in $D^{t+1}$. Similarly,  $p_j$ is the linear index of the paragraph containing $x_j^{t,g}$, and $\#P^{t}$ is the total number of paragraphs in $D^{t}$.
 For example, a sentence from the conclusion is more likely aligned with one from the final parts rather than the introduction. 
 \item If the candidate list is empty after step 3, the sentence $X_i^{t+1,g}$ stays unaligned, indicating its addition. Finally, if a sentence in $D^t$ ends up unaligned to any in $D^{t+1}$, it is pre-annotated as a deletion. 
\end{enumerate}
The similarity thresholds \textit{t0}, and \textit{t1} are optimized in a pilot study on 20 document pairs, where the ideal configuration was determined to be \textit{t0}=40, and \textit{t1}=85, with a similarity of 100 indicating a perfect match.

\subsection{Cross-document Annotation Interface}
\label{subsec:appendix_inception}

For the human annotation process, we utilized the INCEpTION platform \citep{inception}. We developed a \textbf{cross-document environment}\footnote{\url{https://github.com/inception-project/inception/tree/main/inception/inception-io-intertext}} 
that offers the complete context of two documents, facilitating full-document revision analysis and various cross-document annotation tasks. Figure \ref{fig:inception} illustrates the annotation interface.

We posit that presenting only two isolated sentences without full document context is insufficient for thorough long document revision annotation. In long papers, crucial content often recurs in sections like the abstract, introduction, and conclusion, making document structure and context essential for accurate revision alignment. Context also plays a significant role in analyzing revision intent. For instance, if the authors change the name of their proposed method, annotators might perceive it as a different method and label it as a semantic change when only given two sentences. However, with the full document context, annotators can recognize a consistent name change throughout the paper, understanding that the referred method remains the same, thus categorizing it as a change for clarity. 

\subsection{Iterative IAA Assessment}
\label{subsec:appendix_iaa}
Table \ref{tab:iaa1} demonstrates that the IAA has progressively improved after implementing iterative quality management between data packages, thereby evidencing
the efficacy of the employed strategy.
\begin{table}[ht]
\tabcolsep=0.08cm
\begin{tabular}[t]{@{}lllll@{}}
\toprule
\begin{tabular}[c]{@{}l@{}}Tasks $\downarrow$ /\\+Data packages $\rightarrow$\end{tabular}
 & 
\begin{tabular}[c]{@{}l@{}}Val.set\end{tabular}  & \begin{tabular}[c]{@{}l@{}}+DP1\end{tabular} & 
\begin{tabular}[c]{@{}l@{}}+DP2\end{tabular} & 
\begin{tabular}[c]{@{}l@{}}+DP3\end{tabular}\\ 
\midrule
S label  & 0.40  & 0.75 & 0.77  & \textbf{0.78} \\
S align   & 0.99  & 1 & 1  & \textbf{1} \\ 
 \bottomrule
\end{tabular}
\caption[IAA]{Inter-annotator agreement measured by Krippendorf's  $\alpha$ on accumulative data packages, which are improved through iterative quality management. S align: sentence edit alignment, S label: sentence edit labeling.}
\label{tab:iaa1}
\end{table}

\begin{figure*}[ht]
\centering
     \begin{subfigure}[b]{0.46\textwidth}
         \centering
         \includegraphics[width=\textwidth]{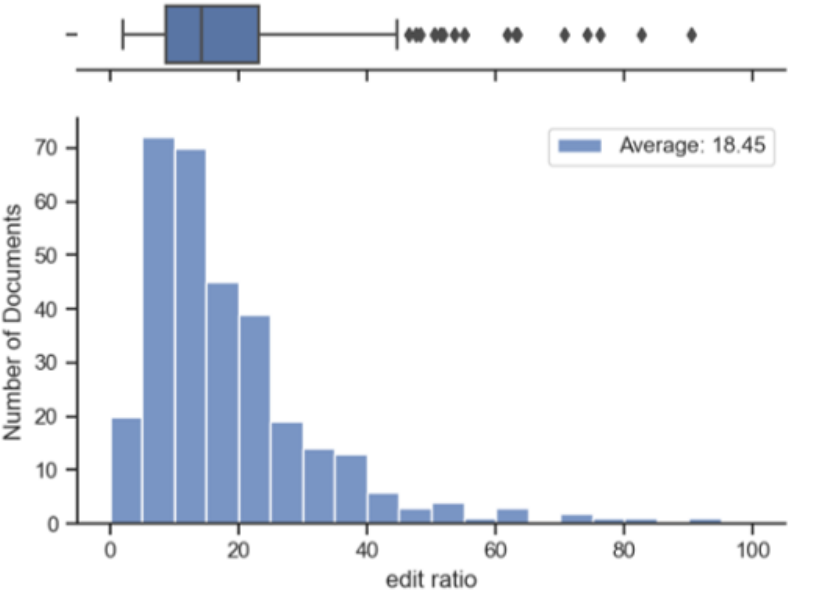}
         \label{fig:edit_ratio}
     \end{subfigure}
     \hfill
     \begin{subfigure}[b]{0.46\textwidth}
         \centering
         \includegraphics[width=\textwidth]{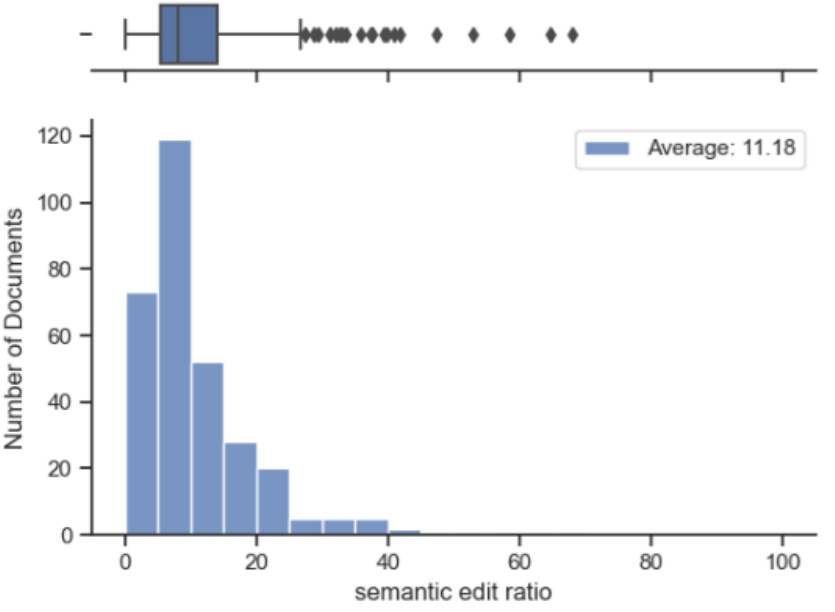}
         \label{fig:semantic_edit_ratio}
     \end{subfigure}
  \caption{Document edit ratio and semantic edit ratio (\%). The y-axis denotes the number of documents.}
  \label{fig:edit_ratio_2}
\end{figure*}

\begin{figure}[ht]
\centering
\includegraphics[width=0.46\textwidth]{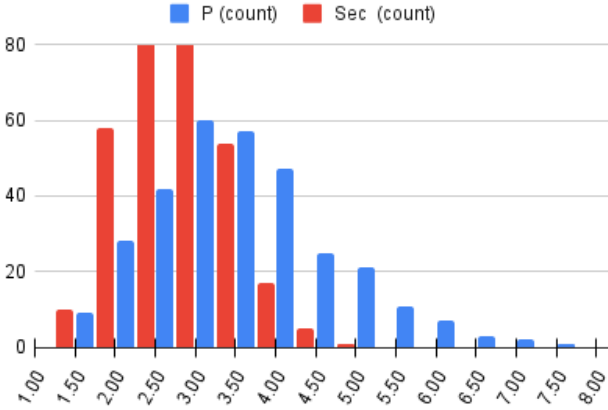}
  \caption{The Crest Factor (CF) of the documents calculated at both paragraph and section levels (P/Sec). The x-axis represents the CF value, while the y-axis shows the count of documents. The CF is a measure of the peak amplitude in a distribution, with an even distribution corresponding to CF=1. For each document, the CF is determined using a vector that denotes the count of sentence edits in each paragraph or section. For example, [0, 0, 0, 0, 0, 0, 0, 0, 0, 0, 2, 12, 0, 0, 0, 0], CF=3.95}
  \label{fig:cf}
\end{figure}

\subsection{Annotators and Tasks}
\label{subsec:annotators}
For the development of the \textbf{annotation environment, taxonomy, and guidelines}, we recruited three annotators: one with expertise in computer science and two specializing in linguistics. The components were iteratively refined based on the annotators' feedback, and further validated in a pilot study with 20 documents, ensuring the robustness and applicability of our methodologies in a practical setting.

For the \textbf{subsequent annotation}, six master's students possessing C1-level English proficiency were recruited, including two in-house annotators who contributed to the prior development. Among these annotators, four specialize in linguistics, one has a background in computer science and one in engineering.
For the sentence-level edit alignment and labeling tasks, each sample is annotated by three annotators, including one in-house annotator among them. A perfect IAA of 1 Krippendorff's $\alpha$ for alignment and a substantial IAA of 0.78 $\alpha$ for labeling was achieved (§\ref{subsec:appendix_iaa}). 
For tasks at the subsentence level, a linguistics expert verified and annotated the edit spans and alignments, which were subsequently labeled for action and intent by three annotators, resulting in an IAA of 0.76 $\alpha$. 
In the tasks of extracting and labeling review requests (refer to RQ5 in §\ref{sec:analysis}), eight documents were annotated by two annotators, achieving an IAA of 0.75 $\alpha$.
Given the substantial IAA, the remaining documents were evenly distributed and annotated individually by each annotator. 
Likewise, the task of associating review requests with their corresponding revisions was undertaken by two annotators on eight documents, resulting in an IAA of 0.68 $\alpha$. Subsequent documents were distributed and each was handled by a single annotator.
Each of the six annotators participated in generating edit summaries, with one annotator assigned per document. They then linked the summary sentences to the corresponding edits.

The \textbf{human evaluation} of the LLM-generated document edit summaries was conducted by the first author of this study, an NLP researcher with expertise in both linguistics and computer science. This researcher also assessed the factuality and specificity of the human-authored summaries (§\ref{subsec:task_summ}).

\subsection{Annotator Survey}
\label{subsec:appendix_survey}
\begin{table*}[ht]
\tabcolsep=0.1cm
\begin{tabular}{ll}
\toprule
Survey Questions & Avg. Score\\\hline 
\textbf{Guideline} \\\hline
 The annotation tasks in the guideline are clear to me.   & 6.8 \\
 The label schema, the definitions and the examples in the guideline are clear to me. & 6.4 \\
 The \textit{annotation procedure} in the guideline is clear to me. & 6.8 \\ 
\begin{tabular}[c]{@{}l@{}}
With the guideline and the annotation demos, I know how to use the annotation tool \\to accomplish the annotation tasks. \end{tabular}  & 7.0 \\ \hline
 \textbf{Training} \\\hline
 \begin{tabular}[c]{@{}l@{}}
 The test annotation, discussion and correction on the first validation set have \\improved my understanding of the tasks and the labels.\end{tabular} & 6.2  \\
\begin{tabular}[c]{@{}l@{}}
 Discussions on my individual questions have improved my understanding of the \\tasks and the labels. \end{tabular} & 6.8 \\
 \begin{tabular}[c]{@{}l@{}}
 Discussions on the summarized common mistakes have improved my understanding \\of the tasks and the labels.\end{tabular} & 6.8  \\
 \begin{tabular}[c]{@{}l@{}}
 Over time, my uncertain samples and questions have decreased significantly.  \end{tabular} & 6.8  \\ \hline
 \textbf{Label taxonomy} \\\hline
 The edit action labels can completely cover the edit actions seen. & 6.8 \\
 The edit intent labels can completely cover the edit intents seen. & 6.2 \\ 
 \hline
 \textbf{Cross-document annotation environment} \\\hline
 For \textit{edit alignment and action labeling}, the cross-document context is crucial. & 6.8 \\
 For \textit{edit intent labeling}, the cross-document context is crucial. & 6.4 \\ \hline
 \textbf{Challenge} \\\hline
It is hard to detect alignment and label the edit action. & 2.2 \\
It is hard to detect the actual differences of a revision pair. & 3.8 \\
It is hard to label the edit intent. & 4.0 \\ \hline
 \textbf{Usability of the annotation tool} \\\hline
 The tool's capabilities meet my requirements. & 5.8 \\
 Using the tool is a frustrating experience. & 2.8 \\
 The tool is easy to use. & 5.2 \\
 I have to spend too much time correcting things with the tool. & 1.8 \\ 
 Average UMUX score & $77\pm15$ \\ \hline
\end{tabular} 
\caption[annotator_survey]{Annotator survey. Annotators are presented with the statements and are asked to rate their level of agreement or disagreement on a 7-point scale, where 1 represents 'Strongly Disagree' and 7 represents 'Strongly Agree'. The final section displays a UMUX survey to measure the usability of the annotation tool and the average system UMUX score.}
\label{tab:annotator_survey}
\end{table*}

Following annotation, annotators complete a survey evaluating the effectiveness and efficiency of the guideline, training process, label taxonomy, cross-document annotation environment, and the annotation tool's usability. They are presented with various statements and asked to rate their agreement or disagreement on a 7-point scale, where 1 signifies 'Strongly Disagree' and 7 'Strongly Agree'.

Table \ref{tab:annotator_survey} indicates that annotators find the guidelines, label taxonomy, and annotation interface adequate, with the iterative training process being particularly effective. They highlight the value of discussions on individual queries and common mistakes as the most beneficial aspect of the training. 
Regarding the taxonomy, annotators report that the existing taxonomy adequately encompasses all observed revisions, and they did not feel the need for additional labels during the annotation process.
Additionally, they recognize the importance of the cross-document annotation environment, especially in aligning revision pairs and labeling edit actions. This is also evident in their assessments of the subtasks' difficulty, with the greatest challenges being change detection and intent identification. Moreover, they perceive the annotation tool as highly usable, as indicated by the UMUX \cite{umux} survey and the average system UMUX score.\footnote{The average system UMUX score is calculated according to: \href{https://blucado.com/understanding-the-umux-a-guide-to-the-short-but-accurate-questionnaire/}{https://blucado.com/understanding-the-umux-a-guide-to-the-short-but-accurate-questionnaire/}} 

\section{Dataset analysis}
\label{sec:appendix_c}
Table \ref{tab:ea_ei_dist} demonstrates the proportions of each edit action or intent label. 
Figure \ref{fig:edit_ratio_2} illustrates the distribution of both the document edit ratio and the semantic document edit ratio. 
Figure \ref{fig:cf} displays the distribution of the Crest Factor (CF) for the documents, measured at the paragraph and section levels.

\begin{table}[ht]
\tabcolsep=0.1cm
\begin{subtable}[h]{0.4\textwidth}
\begin{tabular}{llllll}
\toprule
  Add & Delete & Modify & Merge & Split & Fusion \\ \hline 
  28.93 & 12.44 & \textbf{54.54} & 1.53 & 2.3 & 0.26 \\ \bottomrule
\end{tabular}
\caption{Proportions (\%) of each edit action label.}
\label{tab:ea_dist}
\end{subtable}
\begin{subtable}[h]{0.45\textwidth}
\begin{tabular}{lllll}
\toprule
 Grammar & Clarity & Fact/Evidence & Claim & Other \\ \hline 
 14.38 & 21.78 & \textbf{45.02} & 15.44 & 2.68 \\ \bottomrule
\end{tabular}
\caption{Proportions (\%) of each edit intent label.}
\label{tab:ei_dist}
\end{subtable}
\caption{Edit action and intent distributions.}
\label{tab:ea_ei_dist}
\end{table}

\begin{table}[ht]
\fontsize{10}{10}
\selectfont
\renewcommand{\arraystretch}{1.2} 
\begin{tabular}[t]{ll|ll}
\hline
\multicolumn{4}{c}{\textbf{Baselines}} \\ \hline
Random   & 0.34 \\
diff3-maj & 0.73  & cat3-maj & 0.72    \\
diff5-maj  & 0.75  & cat5-maj & 0.74    \\
diff8-maj & \underline{0.76}  & cat8-maj & \underline{0.76}    \\
\hline
\multicolumn{4}{c}{\textbf{Our Models}} \\ \hline
\multicolumn{2}{l}{\textit{ICL \& CoT}} & \multicolumn{2}{l}{} \\ \hline
\multicolumn{4}{l}{\textcircled{\tiny 1}\textit{+ dynamic examples}}  \\ 
+diff3 & \underline{0.74}  & +cat3  & 0.73    \\ \hline
\multicolumn{4}{l}{\textcircled{\tiny 2}\textit{+ static examples}}  \\ 
 \begin{tabular}{l}
     +def3-(L,R)
\end{tabular} & 0.79 &&  \\ \hline
\multicolumn{4}{l}{\textcircled{\tiny 3}\textit{+ def3-(L,R) + dynamic }}   \\ 
+diff3 & \underline{\textbf{0.83}}  &   \\ \hline
\multicolumn{4}{l}{\textcircled{\tiny 4}\textit{+ dynamic + def3-(L,R)}} \\ 
+diff3 & \underline{\textbf{0.83}} &+cat3 &0.80  \\
+diff5 &0.82 &+cat5 &0.80  \\
\hline
\end{tabular}
\caption[LLMs]{Llama2-70B accuracy in edit intent classification on addition and deletion samples. Baselines are assessed on the full test set, subsequent models are evaluated on 20\% of the test set for validation. Scores underlined represent the best accuracy within the same method block, with the highest accuracy in bold. The numbers in the model names indicate the number of selected demonstrations. As detailed in Section \ref{sec:analysis}, since additions and deletions are exclusively associated with Fact/Evidence, Claim, and Other, we use three default examples.}
\label{tab:ad_intent}
\end{table}

\begin{table}[ht]
\fontsize{10}{10}
\selectfont
\renewcommand{\arraystretch}{1.2} 
\begin{tabular}[t]{ll|ll}
\hline
\multicolumn{4}{c}{\textbf{Baselines}} \\ \hline
Random   & 0.50  \\
diff3-maj & 0.59  & cat3-maj & \underline{0.74}    \\
diff5-maj  & 0.58  & cat5-maj & 0.73    \\
\hline
\multicolumn{4}{c}{\textbf{Our Models (ICL \& CoT)}} \\ \hline
\multicolumn{4}{l}{\textcircled{\tiny 1}\textit{+ dynamic examples}}  \\ 
+diff3 & \underline{0.95}  & +cat3  & 0.94    \\ \hline
\multicolumn{4}{l}{\textcircled{\tiny 2}\textit{+ static examples}}  \\ 
 \begin{tabular}{l}+def2-(L,R)
\end{tabular} & \underline{\textbf{0.97}} & \begin{tabular}{l}+def2-(R,L)
\end{tabular} & 0.95  \\ \hline
\multicolumn{4}{l}{\textcircled{\tiny 3}\textit{+ def2-(L,R) + dynamic} }   \\ 
+diff3 & 0.96  &   \\ \hline
\multicolumn{4}{l}{\textcircled{\tiny 4}\textit{+ dynamic + def2-(L,R)}} \\ 
+diff3 & \underline{\textbf{0.97}} &+cat3 &\underline{\textbf{0.97}}  \\
+diff5 &0.96 &+cat5 &\underline{\textbf{0.97}}\\
\hline
\end{tabular}
\caption[alignment]{Llama2-70B accuracy in revision alignment. Baselines are assessed on the full test set, subsequent models are evaluated on 20\% of the test set for validation. Scores underlined represent the best within the same method block, with the highest accuracy in bold. The numbers in the model names indicate the number of selected demonstrations.}
\label{tab:revision_align}
\end{table}

\section{Automation with LLMs}
\label{sec:appendix_d}
\subsection{Edit intent classification}
\label{subsec:appendix_task_intent}
Table \ref{tab:prompt} shows an example prompt with system instruction, demonstration and task instruction used for experiments. 
Table \ref{tab:ad_intent} presents the performance of Llama2-70B in identifying edit intents for additions and deletions. Echoing the findings in Table \ref{tab:m_intent} on revision pairs, we observe that default examples with CoT reasoning yield strong results. These outcomes are further enhanced when three demonstrations selected via the \textit{diff} method are included.
Table \ref{tab:adm_intent} presents the results of the joint evaluation conducted on all 8,937 test samples. The challenge in identifying edit intents is particularly evident in revision pairs, highlighted by the low precision in Clarity, low recall in Fact/Evidence, and the difficulties associated with low-sourced Claim and Other classes.

\begin{table*}[ht]
\fontsize{10}{10}
\selectfont
\renewcommand{\arraystretch}{1.2} 
\tabcolsep=0.10cm
\begin{tabular}{l|ll|lll|lll|lll|lll|lll}
\hline
class/ &\multicolumn{2}{c}{Total} &\multicolumn{3}{c}{Grammar}
&\multicolumn{3}{c}{Clarity}&\multicolumn{3}{c}{Fact/Evidence}
&\multicolumn{3}{c}{Claim}&\multicolumn{3}{c}{Other}\\ 
count &\multicolumn{2}{c}{8937}&\multicolumn{3}{c}{1309}
&\multicolumn{3}{c}{1838}&\multicolumn{3}{c}{4110}
&\multicolumn{3}{c}{1432}&\multicolumn{3}{c}{248}\\ \hline
metrics & Acc. & M. F1  & P & R & F1  & P & R & F1
& P & R & F1
 & P & R & F1 & P & R & F1\\ \hline
\multicolumn{18}{c}{\textbf{Baselines}} \\ \hline
Random   & 0.20 &  0.18 
&0.15&0.20&0.17  &0.21&0.20&0.20 
&0.46&0.20&0.28&0.17&0.21&0.19&0.03&0.21&0.05 \\
Majority  & 0.46 & 0.13
 &0&0&0  &0&0&0 &0.46&1&0.63&0&0&0&0&0&0\\
\hline
\multicolumn{18}{c}{\textbf{Our Model (ICL\&CoT)}} \\ \hline
\multicolumn{18}{l}{\textit{+ diff3 + def-(L,R)}} \\ 
joint & 0.7 & 0.69  
& 0.79 & 0.72 & 0.75
& 0.54 & 0.85 & 0.66 
& 0.85 & 0.68 & 0.76
& 0.61 & 0.58 & 0.6
& 0.76 & 0.62 & 0.69 \\\hline
A,D/ count &\multicolumn{2}{c}{3891} 
&\multicolumn{3}{c}{0}&\multicolumn{3}{c}{0}
&\multicolumn{3}{c}{2580}&\multicolumn{3}{c}{1135}
&\multicolumn{3}{c}{176}\\
&0.78 & 0.77 & - & - & -& - & - & -
& 0.86 & 0.8 & 0.83 &0.62&0.73&0.67&0.85&0.76&0.80 \\\hline
R/ count &\multicolumn{2}{c}{5046} 
&\multicolumn{3}{c}{1309}&\multicolumn{3}{c}{1838}
&\multicolumn{3}{c}{1530}&\multicolumn{3}{c}{297}
&\multicolumn{3}{c}{72}\\
&0.65 & 0.48 & 0.79 & 0.72 & 0.75
& 0.54 & 0.85 & 0.66
& 0.82 & 0.48 & 0.61
& 0.14 & 0.01 & 0.01
&0.46 & 0.29 &0.36 \\\hline
\end{tabular}
\caption[joint_intent]{Edit intent classification,  joint evaluation of the optimal configuration using Llama2-70B on all test samples. R: revision pairs, A,D: additions and deletions. Displayed are the accuracy (Acc.), macro average F1 score (M. F1), and precision (P), recall (R), and F1 score for each label. The challenge in identifying edit intents is particularly evident in revision pairs, highlighted by the low precision in Clarity, low recall in Fact/Evidence, and the difficulties associated with low-sourced Claim and Other classes.}
\label{tab:adm_intent}
\end{table*}

\begin{table*}[ht]
\fontsize{10}{10}
\selectfont
\renewcommand{\arraystretch}{1.2} 
\tabcolsep=0.10cm
\begin{tabular}{lll}
\toprule
System instruction: & & \\ \hline
\multicolumn{3}{l}{\begin{tabular}{l}
You are a helpful, respectful and honest revision analysis assistant.
You will read two versions of texts.\\ Your task is to analyze the revision intent behind the difference between the two texts. The intent can be \\one of the following labels: fix grammar (Grammar), improve clarity (Clarity), change claim or statement \\(Claim), change factual information (Fact/Evidence). Grammar and Clarity are more about surface \\language improvements, while Fact/Evidence and Claim are more about meaning changes. If none of the \\above labels are relevant, please answer with 'Other'.
\end{tabular}}\\\hline
Demonstration with gold label and CoT rationale: & & \\\hline
\multicolumn{3}{l}{\begin{tabular}{l}
The old text is: Empirical studies on the datasets across 7 different languages confirm the effectiveness of \\the proposed model. \\
The new text is: Empirical studies on the three datasets across 7 different languages confirm the \\effectiveness of the proposed model. \\
LABEL: Fact/Evidence\\
REASON: "Three" is added to the new text. This is an addition of factual information that the empirical \\studies are conducted on "three" datasets, thus the label is Fact/Evidence.
\end{tabular}}\\ \hline
Task instruction: & & \\\hline
\multicolumn{3}{l}{\begin{tabular}{l}
Read the following old and new texts. 
What is the intent of the revision? Please answer with one of the \\labels: Grammar, Clarity, Claim, Fact/Evidence and Other. 
Please always answer with the template and fill\\ the template with your answer without additional texts: LABEL:<your answer> REASON:<your answer>.
\end{tabular}}\\ \bottomrule
\end{tabular}
\caption{Example of a Llama2 prompt for edit intent classification: The prompt encompasses the system instruction,  demonstration examples with gold labels and optional CoT rationale, as well as the task instruction.}
\label{tab:prompt}
\end{table*}

\subsection{Document Edit Summarization}
\label{subsec:appendix_edit_summ}
Table \ref{tab:edit_summ_example} provides examples of human-written and GPT-4 generated document edit summaries.

\subsection{Revision Alignment}
\label{subsec:appendix_align}
Table \ref{tab:revision_align} presents the experimental results in revision alignment. 

\begin{table*}[ht]
\fontsize{10}{10}
\selectfont
\renewcommand{\arraystretch}{1.2} 
\tabcolsep=0.10cm
\begin{tabular}{l}
\toprule
human-written summary\\\hline
\begin{tabular}[c]{@{}l@{}}In the Late complications section, a claim was modified to add a reason for excessive scarring and a sentence was \\changed for clarity.\\
The Comments have an added claim about level I evidence not being available and another one regarding the design \\of RCT being unethical. Three sentences were merged together, with the claim change that the studies should be \\taken into consideration even if they are imprecise and unreliable, as well as a few clarity changes. \\Third-party and author's facts about data collection were removed, and a lot of claims regarding the low reliability of \\data in the studied context were also removed.\\
In the Religious and cultural views section, claims about gender bias were excluded. \\Finally, the Final remarks section title was changed to Conclusions.\end{tabular} 
\\ \hline
GPT-4 generated summary\\\hline
\begin{tabular}[c]{@{}l@{}}
The revisions made to the long document can be summarized as follows:\\
1. Deletions:
   \\- Several claims and statements regarding the testing and approval of new drugs, surgical procedures, \\and the comparison of procedures were deleted from the "Comments" section.
   \\- A claim about the gender-based nature of the ban against Female Genital Cutting (FGC) and a suggestion for \\anti-FGC groups to advise on making FGC as safe as male circumcision were removed from \\the "Religious and cultural views" section.
   \\\colorbox{pink}{- A statement about the unreliability of research based on reported data and individual memory was deleted} \\\colorbox{pink}{from the "Comments" section.}
   \\- A claim about the strongest evidence coming from randomized controlled trials and cohort studies was removed \\from the "Comments" section.\\
2. Additions:\\
   - Two claims were added to the "Comments" section, one stating that level I evidence to support the ban against FGC \\is not available and another mentioning the ethical issues in designing a randomized controlled trial (RCT) to address \\the effects of FGC.\\
3. Modifications:\\
   - The title of the "Female circumcision: Limiting the harm" section was changed from "Final remarks" to "Conclusions."
   \\- A claim in the "Late complications" section was modified to include additional details about the possible causes of \\increased Cesarean deliveries in cut women.
   \\- Another statement in the "Late complications" section was modified for clarity, emphasizing that the high Cesarean rate\\ cannot be solely attributed to one cause.
   \\- In the "Comments" section, a merge and modification were made to combine several statements about the evidence\\ supporting the practice of female circumcision, highlighting the consideration of level III evidence despite its imprecision \\and low reliability.\\
Overall, the revisions involved removing certain claims and evidence that may have been unsupported or contentious, \\adding statements about the lack of high-level evidence against FGC, and modifying some sections for clarity and \\to provide a more nuanced view of the evidence available.
\end{tabular} 
\\ \hline
\end{tabular}
\caption[]{Examples of human-written and GPT-4 generated document edit summaries. The summary sentence in pink is annotated as incorrect - factual information instead of a statement was removed.}
\label{tab:edit_summ_example}
\end{table*}

\subsection{Computational details}
\label{subsec:appendix_task_gpu}
In our classification tasks with Llama2-70B, covering revision alignment, edit intent classification, and review request extraction, we employed two RTX™ A6000 GPUs, each equipped with 48GB of memory. The batch size for inference was established at four.
For the document edit summarization task using GPT-4, we processed 282,964 input tokens and produced 36,341 output tokens in total, resulting in a total expense of 3.92 US dollars.

\begin{table*}[ht]
\fontsize{10}{10}
\selectfont
\renewcommand{\arraystretch}{1.2} 
\tabcolsep=0.10cm
\begin{tabular}{lll}
\toprule Example &\begin{tabular}[c]{@{}l@{}}Revision description\end{tabular} 
& Notation \\\hline 
  \begin{tabular}[c]{@{}l@{}}These findings constitute the first \\ evidence that using our taxonomy \\ could result in robust methods\colorbox{lightblue}{, even} \\ \colorbox{lightblue}{though more data and research} \\ \colorbox{lightblue}{seem necessary to get there}.\end{tabular}  & \begin{tabular}[c]{@{}l@{}}A \textbf{subsentential} text element, i.e., the \\highlighted  clause, is \textit{added} for a more \\cautious view,  \textit{claiming} that further \\ data and research are  required. If viewed \\at the \textbf{sentence} level,  this reflects a   \\\textit{modification} of an existing sentence. \end{tabular} &\begin{tabular}[c]{@{}l@{}}(\textbf{SS}, \textit{Add, Claim}) \\or\\
  (\textbf{S}, \textit{Modify, Claim})\end{tabular}\\ \hline
  
  \begin{tabular}[c]{@{}l@{}}... is a medication for smoke \\cessation. \colorbox{lightblue}{All these cases pose } \\ \colorbox{lightblue}{challenges to state-of-the-art } \\ \colorbox{lightblue}{language models.} Recent work ...\end{tabular} &\begin{tabular}[c]{@{}l@{}}An entire new \textbf{sentence} is \textit{added} to \\make a \textit{claim}. If viewed at the \\\textbf{paragraph} level, this represents a \\\textit{modification} of an existing paragraph. \end{tabular} & \begin{tabular}[c]{@{}l@{}}(\textbf{S}, \textit{Add, Claim})\\or\\(\textbf{P}, \textit{Modify, Claim})\end{tabular}\\ \hline
  
  \begin{tabular}[c]{@{}l@{}} \colorbox{lightblue}{ 
 The values were compared using } \\\colorbox{lightblue}{the Bonferroni test post hoc. Also,}\\ \colorbox{lightblue}{the population density of each } \\\colorbox{lightblue}{zone was calculated ... }\end{tabular}  & \begin{tabular}[c]{@{}l@{}}\textit{Addition} of one entire new \textbf{paragraph} \\to furnish \textit{factual details}. From a \\ \textbf{sentence-level} view, this equates to\\ multiple sentence \textit{additions}.\end{tabular}& \begin{tabular}[c]{@{}l@{}}(\textbf{P}, \textit{Add, Fact/Evidence})\\or multiple \\ (\textbf{S}, \textit{Add, Fact/Evidence})\end{tabular} \\ \hline
 
  \begin{tabular}[c]{@{}l@{}} 
 However, the problem is that the \\ hypothesis has limitations in reflecting a\\ word's meanings\st{.}\colorbox{lightblue}{, because w}\st{W}ords \\having different  \colorbox{lightblue}{or even opposite} \\meanings can appear in similar contexts. \end{tabular}  & \begin{tabular}[c]{@{}l@{}}An existing \textbf{paragraph} is \textit{modified} to \\update claims. Upon closer inspection at \\ the \textbf{sentence}  level, it involves \textit{merging} \\two sentences for \textit{clarity} and \textit{modifying} \\ one of them to update \textit{claims}. \end{tabular}& \begin{tabular}[c]{@{}l@{}}(\textbf{P}, \textit{Modify, [Clarity, Claim]})\\or\\(\textbf{S}, \textit{Merge+Identical, Clarity})\\(\textbf{S}, \textit{Merge+Modify, Claim})\end{tabular}\\ \hline
\end{tabular} 
\caption[edit intents]{Revision examples described and notated by the three revision dimensions: granularity, action and intent. The same revisions are perceived differently based on varying levels of granularity, making the three dimensions necessary for a precise analysis. Texts with strikethroughs are removed, and texts highlighted in blue are added. SS: subsentence-level, S: sentence-level, P: paragraph-level.}
\label{tab:3_dim_examples}
\end{table*}

\begin{table*}[]

\tabcolsep=0.10cm
\begin{tabular}{llll}
\toprule
\multicolumn{2}{l}{Edit actions} & Definitions  & Alignment type\\\hline  
Content & Add  & \begin{tabular}[c]{@{}l@{}}Insert an entire new text element \end{tabular}& 1-to-0 \\\cline{2-4}
&Delete& \begin{tabular}[c]{@{}l@{}}Remove an existing text element completely\end{tabular} &0-to-1\\ \cline{2-4}
&Modify & \begin{tabular}[c]{@{}l@{}}Revise an existing text element by altering a portion of it, with \\some parts of the original text remaining unchanged.\end{tabular}  & 1-to-1 \\ \hline
Partition & Merge& \begin{tabular}[c]{@{}l@{}}Consolidate multiple text elements into a single text element\end{tabular}&1-to-n\\\cline{2-4}
& Split& \begin{tabular}[c]{@{}l@{}}Distribute a single text element into multiple separate text elements\end{tabular}  & n-to-1\\ \cline{2-4}
& Fusion& \begin{tabular}[c]{@{}l@{}}Combination of merge(s) and split(s) \end{tabular}  & m-to-n\\ \hline
\end{tabular} 
\caption[edit action]{Edit action definitions and alignment types. The alignment refers to the new-version-to-old-version relation, the new version is the source of the alignment and the old version is the target. Partition changes comprise a series of one-to-one alignments, each necessitating an accompanying content label (Modify or Identical) to denote whether the linked element's content has altered (also see the last example in Table \ref{tab:3_dim_examples}).
 }
\label{tab:edit_actions}
\end{table*}

\begin{table*}[]
\fontsize{10}{10}
\selectfont
\renewcommand{\arraystretch}{1.2} 
\tabcolsep=0.10cm
\begin{tabular}{lllll}
\toprule
\multicolumn{2}{l}{Edit intents} & Definitions  & Subsentence-level Examples \\
\hline  \multicolumn{1}{l}{Surface} & Grammar  & \begin{tabular}[l]{@{}l@{}}Correct grammatical errors, capitalization, \\ punctuation, tense, modality, spelling, \\typography,  abbreviations or any errors \\ related to grammar and/or conventions to \\ improve the language.\end{tabular} & \begin{tabular}[c]{@{}l@{}} \textbf{Modify, Grammar:}\\It is freely available for\\ \st{akademic}\colorbox{lightblue}{academic} use.\end{tabular}\\ \cline{2-4}
  & Clarity       & \begin{tabular}[c]{@{}l@{}}Alter word choice, phrase usage, expressions \\ and/or text format to be more formal, concise \\ and understandable without meaning changes,\\ or to amplify meaning for clarity.\end{tabular} & \begin{tabular}[c]{@{}l@{}}\textbf{Modify, Clarity:}\\This study \st{checked out}\colorbox{lightblue}{examined} \\ how images affect learning.\end{tabular} \\ \hline
\multicolumn{1}{l}{Semantic} &Fact/Evidence & \begin{tabular}[c]{@{}l@{}}Add, elaborate, extend, verify or update the \\ fact and/or evidence from third parties, or \\ the author's factual manipulations and \\observations, or delete/modify \\erroneous/irrelevant ones.\end{tabular}  & \begin{tabular}[c]{@{}l@{}}\textbf{Modify, Fact/Evidence: }\\XX, et. al. sets the state-of-the-art\\ ROUGE result to \st{0.56}\colorbox{lightblue}{0.54}.\end{tabular} \\ \cline{2-4}
  \multicolumn{1}{l}{} & Claim         & \begin{tabular}[c]{@{}l@{}}Change/Add/Delete the claim, statement,\\  opinion,  idea of the authors, or their overall \\ aim of the document.\end{tabular}  & \begin{tabular}[c]{@{}l@{}}\textbf{Add, Claim:}\\These findings constitute the first \\ evidence that using our taxonomy \\ could result in robust methods\colorbox{lightblue}{, even} \\ \colorbox{lightblue}{though more data and research} \\ \colorbox{lightblue}{seem necessary to get there}.\end{tabular} \\ \hline
\multicolumn{1}{l}{Other} &  & \begin{tabular}[c]{@{}l@{}}Revise the text in a way that is irrelevant \\ to any other type of edit intent.\end{tabular}  
& \begin{tabular}[c]{@{}l@{}}\textbf{Add, Other} (\textit{changes in section titles}): \\ Experiments \colorbox{lightblue}{and Results}\end{tabular}  \\ \hline
\end{tabular} 
\caption[edit intents]{Edit intent definitions and subsentence-level examples. Texts with strikethroughs are removed, and texts highlighted in blue are added.}
\label{tab:edit_intents}
\end{table*}

\end{document}